\documentclass[11pt]{article}
\pdftrailerid{}
\usepackage{fullpage}
\usepackage{algorithmic}
\usepackage{algorithm}
\usepackage{graphicx,subcaption} %
\usepackage{amsfonts,amssymb,amsmath,amsthm,amsopn}	%
\usepackage{hhline,booktabs,colortbl,multirow,tabularx,threeparttable} %

\usepackage{enumerate}
\usepackage{footnote}
\usepackage{float}
\usepackage{hyperref}
\usepackage{prettyref}
\usepackage{setspace}
\usepackage{color}
\usepackage{xcolor}  %
\usepackage{titlesec} %
\usepackage{reporttitleblock} %
\usepackage{geometry}

\usepackage{tikz}
\usepackage{pgfplots}
\usetikzlibrary{positioning,shapes,shadows,arrows,calc}
\tikzstyle{component}=[rectangle, draw=black, rounded corners, fill=blue!40, drop shadow, text centered, anchor=north, text=white, minimum height=1cm]
\tikzstyle{arrow}=[->, thick]

\pgfplotsset{compat=1.18}
\usetikzlibrary{intersections}
\usetikzlibrary{pgfplots.statistics}
\usepgfplotslibrary{fillbetween}

\usepackage{eso-pic}   %
\usepackage{titling}   %
\usepackage{fancyhdr}  %
\usepackage{lastpage}  %
\usepackage{fontawesome5} %

\newcommand{\PageCounterText}{%
  {\small\color{headerblue}\faBookOpen~~\textbf{\thepage}/\pageref{LastPage}}%
}
\newcommand{\FirstPageFooterLeft}{%
  \small{\color{headerblue}\faBook}~~%
  {\small\color{headerblue}\textbf{\ReportTitleMarkText\ | October 13, 2026}}%
}
\fancypagestyle{allpagesfooter}{%
  \fancyhf{}%
  \fancyfoot[R]{\PageCounterText}%
}
\fancypagestyle{firstpagefooter}{%
  \fancyhf{}%
  \fancyfoot[L]{\FirstPageFooterLeft}%
  \fancyfoot[R]{\PageCounterText}%
}
\definecolor{myblue}{RGB}{34,31,217}
\definecolor{mycyan}{gray}{.7}
\definecolor{Gray}{gray}{0.9}
\definecolor{headerblue}{RGB}{0,85,150}    %
\definecolor{natureblue}{RGB}{11,61,145}   %
\definecolor{titlegray}{RGB}{80,80,80}     %
\definecolor{titlebox}{RGB}{245,248,252}   %
\newlength{\abstractpad}
\hypersetup{
  pdftitle={TANGO: RNA Topology and Geometry Co-Design},
  pdfauthor={Tianmeng Hu, Biao Luo, Ke Li},
  colorlinks=true,
  linkcolor=natureblue,
  citecolor=natureblue,
  urlcolor=natureblue,
  filecolor=natureblue
}

\titleformat{\section}{\color{natureblue}\normalfont\Large\bfseries}{\thesection}{1em}{}
\titleformat{\subsection}{\color{natureblue}\normalfont\large\bfseries}{\thesubsection}{1em}{}
\titleformat{\subsubsection}{\color{natureblue}\normalfont\normalsize\bfseries}{\thesubsubsection}{1em}{}

\newcommand{\pref}{\prettyref}

\newrefformat{fig}{Figure~\ref{#1}}
\newrefformat{tab}{Table~\ref{#1}}
\newrefformat{sec}{Section~\ref{#1}}
\newrefformat{app}{Appendix~\ref{#1}}
\newrefformat{alg}{Algorithm~\ref{#1}}
\newrefformat{property}{Property~\ref{#1}}
\newrefformat{theorem}{Theorem~\ref{#1}}
\newrefformat{corollary}{Corollary~\ref{#1}}
\newrefformat{proposition}{Proposition~\ref{#1}}
\newrefformat{def}{Definition~\ref{#1}}
\newrefformat{eq}{Eq.~(\ref{#1})}

\usepackage{lscape}

\usepackage[numbers,compress]{natbib}
\newcommand{\our}{\texttt{TANGO}}

\usepackage[utf8]{inputenc}
\usepackage[T1]{fontenc}
\usepackage{nicefrac}
\usepackage[nopatch=footnote]{microtype}
\usepackage{array}
\usepackage{url}
\newrefformat{ass}{Assumption~\ref{#1}}
\usepackage[section]{placeins}
\usepackage{needspace}
\renewcommand{\topfraction}{0.9}

\renewcommand{\textfraction}{0.08}
\renewcommand{\floatpagefraction}{0.65}

\newcommand{\tangoalgspacing}{\setlength{\itemsep}{0.20em}\setlength{\parsep}{0pt}}


\ReportClearAffils{}
\ReportClearAuthors{}
\ReportTitle{TANGO: RNA Topology and Geometry Co-Design}
\ReportTitleMark{COLAlab Report}
\ReportAcceptedNote{Accepted at NeurIPS 2026.}
\ReportCorrespondingLabel{Corresponding authors.}

\ReportAffil{1}{Earlham Institute, UK.}
\ReportAffil{2}{Living Systems Institute, University of Exeter, UK.}
\ReportAffil{3}{Department of Computer Science, University of Exeter, UK.}
\ReportAffil{4}{School of Automation, Central South University.}
\ReportAffil{5}{School of Computing Sciences, University of East Anglia, UK.}

\ReportAuthor{Tianmeng Hu}{1,2,3}{tianmeng0824@gmail.com}
\ReportAuthor{Biao Luo}{4}[corresponding]{biao.luo@hotmail.com}
\ReportAuthor{Ke Li}{1,2,3,5}[corresponding]{Ke.Li@earlham.ac.uk}

\begin{document}
\thispagestyle{firstpagefooter}

\AddToShipoutPictureBG*{%
  \AtPageUpperLeft{%
    \raisebox{-2.6cm}{\hspace{2cm}%
      \parbox[t]{\textwidth}{%
        \includegraphics[height=1.4cm]{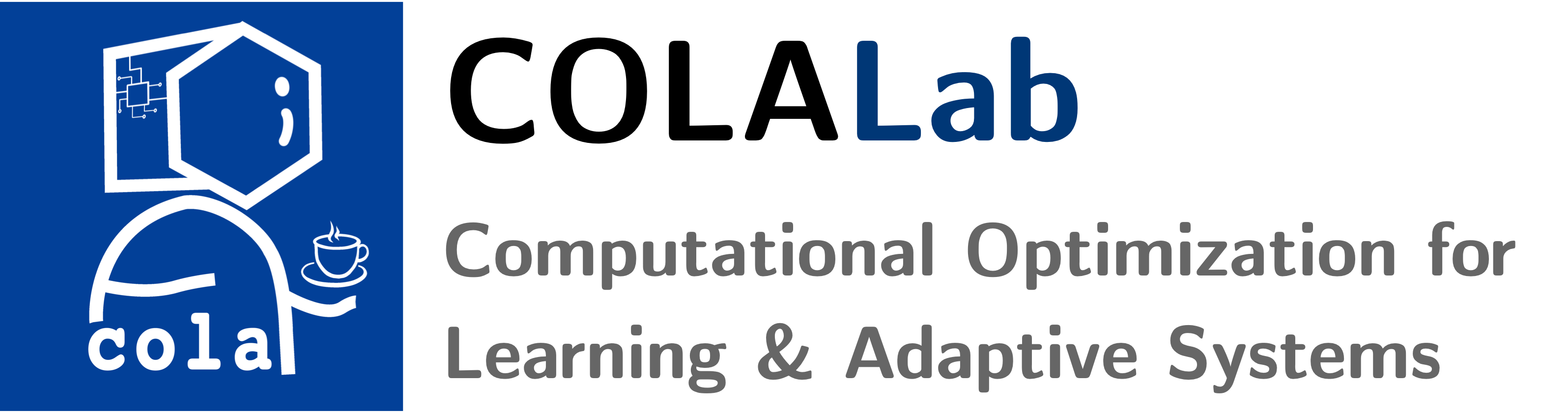}\par
      }%
    }%
  }%
}

\vspace*{0.8cm}

\MakeReportTitleBlock{}

\vspace{1.0em}

\noindent
	\begin{tikzpicture}
	  \node[
	    fill=titlebox,
	    rounded corners=10pt,
	    inner xsep=\abstractpad,
	    inner ysep=10pt
	  ]{%
	    \begin{minipage}{\dimexpr\textwidth-2\abstractpad\relax}
	      \setlength{\parindent}{0pt}%
	      {\bfseries\color{natureblue} Abstract}\par\medskip
Designing RNA sequences that satisfy target tertiary-structure objectives is an important problem in RNA engineering, with broad applications in synthetic biology and therapeutics. Existing $\mathrm{3D}$ RNA design methods have made substantial progress, but they largely treat the target fold as a geometry-only objective. This is incomplete for RNA, whose functional folds are specified not only by tertiary geometry but also by secondary base-pairing topology. We therefore formulate RNA inverse design as a \emph{topology-and-geometry co-design} problem and propose \our, a \emph{multi-objective multi-agent reinforcement learning} framework for this task. \our\ adopts a staged curriculum: it first learns a topology-aware policy from secondary-structure targets, then transfers this policy to tertiary-structure targets to train a general co-design model, and finally applies frontier-guided fine-tuning to refine target-specific trade-offs among secondary-structure fidelity, tertiary-structure fidelity, and sequence diversity. Experiments show that \our\ generates diverse and novel designs, improving tertiary- and secondary-structure performance by $18.3\%$ and $52.7\%$, respectively, over the strongest baseline.

	    \end{minipage}
	  };
	\end{tikzpicture}

\section{Introduction}
\label{sec:introduction}

RNA plays a central role in molecular biology. It participates in every stage of gene regulation, from transcription and processing to export, translation, and degradation~\cite{strobel2007rna,SerganovN13,zhang2022advances}. Unlike DNA, RNA molecules are not static lines
of code. They fold into complex secondary ($\mathrm{2D}$) and tertiary ($\mathrm{3D}$) structures in vivo. These structures are not merely passive geometries; they actively dictate biological traits. Computational RNA inverse design aims to generate sequences that satisfy specified structural objectives~\cite{churkin2018design}, and recent progress in deep learning has made tertiary-structure-based RNA design increasingly practical~\cite{tan2024rdesign,joshi2025grnade,wong2024deep,huang2024ribodiffusion,hu2026rider,ma2026reflex}.

RNA design, however, differs from standard structure-conditioned protein design in a central way: an RNA fold is organized not only by $\mathrm{3D}$ geometry but also by an explicit base-pairing topology. Local Watson-Crick and wobble interactions form stems, loops, junctions, and pseudoknots, providing a secondary-structure scaffold for tertiary assembly~\cite{brion1997hierarchy}. Long-range and non-canonical interactions then stabilize particular $\mathrm{3D}$ conformations within the topologically allowed ensemble \cite{butcher2011molecular, bailor2010topology,mustoe2014topological}. Thus, neither secondary topology nor tertiary geometry alone fully specifies an RNA design objective: topology constrains the feasible fold space, while geometry determines the final spatial arrangement.

Existing $\mathrm{3D}$ RNA design methods have made important progress by conditioning sequence generation on tertiary structures. Some also incorporate secondary-structure information as priors and constraints~\cite{wong2024deep} or repair heuristics~\cite{tan2024rdesign}. Nevertheless, topology fidelity is rarely treated as a first-class optimization and evaluation target jointly with tertiary-structure fidelity. This leaves a gap between what is optimized and what functional RNA design often requires: a generated sequence should not merely fold back to a geometrically similar structure, but should also recover the base-pairing organization that scaffolds that fold. A detailed discussion of related work is provided in~\pref{app:related_work}.

\begin{figure*}[t]
    \centering
    \includegraphics[width=\textwidth]{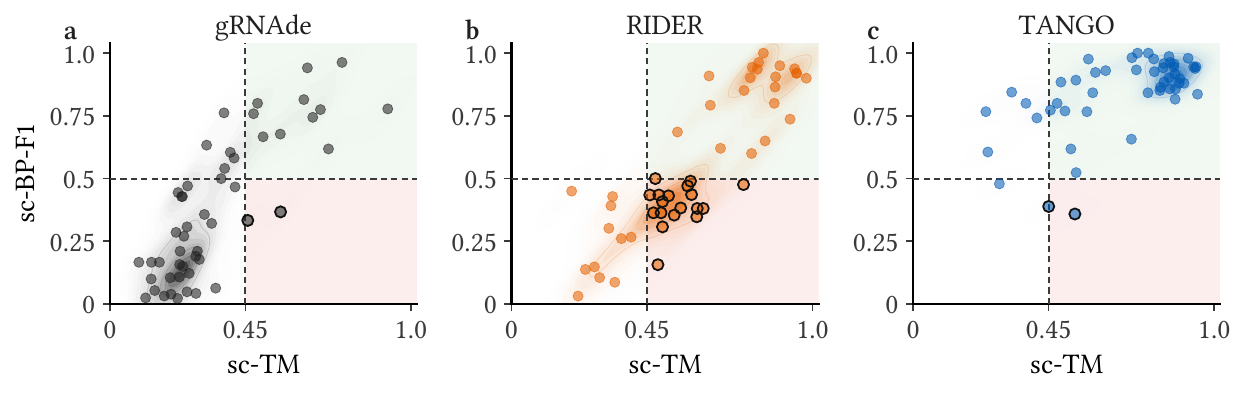}
    \caption{\textbf{RNA $\mathrm{3D}$ inverse design requires joint topology and geometry fidelity.} Each point denotes a designed sequence evaluated by fold-back self-consistency, with tertiary-structure fidelity measured by $\mathrm{sc\text{-}TM}$ and secondary-topology fidelity measured by $\mathrm{sc\text{-}BP\text{-}F1}$. The upper-right region contains designs that satisfy both objectives. Existing $\mathrm{3D}$ RNA design methods can produce geometry-consistent but topology-inconsistent designs, showing that tertiary similarity alone does not guarantee recovery of the target base-pairing topology. Conversely, topology preservation alone is insufficient for specifying the final tertiary fold.}
    \label{fig:intro_failure}
\end{figure*}

This gap is not merely conceptual. As shown in~\pref{fig:intro_failure}, strong tertiary-structure-based baselines can generate candidates with reasonable fold-back geometry but poor recovery of the target base pairs. Conversely, candidates that preserve much of the secondary topology may still fail to realize the desired $\mathrm{3D}$ arrangement. These failure modes indicate that topology fidelity and geometry fidelity are complementary objectives rather than redundant metrics. They also suggest that RNA $\mathrm{3D}$ inverse design should be evaluated as a \textit{co-design problem}: a successful sequence should recover both the target tertiary geometry and the target base-pairing topology.

We therefore formulate RNA $\mathrm{3D}$ inverse design as \emph{topology-and-geometry co-design}. Given a target tertiary structure and its associated base-pairing topology, the goal is to generate sequences whose predicted folds match both. This objective is challenging for two reasons.  First, base-paired positions are combinatorially coupled: changing one nucleotide may require a coordinated change at its pairing partner. Second, tertiary-structure rewards are delayed and sparse, because candidate sequences must be completed and folded before their $\mathrm{3D}$ quality can be assessed. Although the sequence space contains $4^L$ possibilities for length $L$, only a small subset is compatible with a target topology, and only a smaller subset of those sequences realizes the target tertiary geometry.

We propose \our, a \emph{multi-objective multi-agent reinforcement learning} framework for RNA topology-and-geometry co-design. \our\ uses the target topology to decompose sequence generation into structural units: unpaired residues are designed individually, while paired positions are designed jointly. Multiple agents fill these units in parallel under a shared structural reward, reducing the effective decision horizon relative to nucleotide-by-nucleotide generation and enabling coordinated decisions over coupled base pairs. Training proceeds in three stages. First, topology-aware policy initialization learns sequence-construction patterns that satisfy secondary-structure scaffolds. Second, geometry-conditioned training transfers this policy to tertiary-structure targets, allowing $\mathrm{3D}$ information to adjust topology-compatible sequence decisions. Third, frontier-guided fine-tuning adapts the general policy to each target by directly improving the Pareto frontier over secondary-topology fidelity, tertiary-structure fidelity, and sequence diversity. Experiments on a curated PDB-derived benchmark show that \our{} improves tertiary fold-back fidelity by $18.3\%$ and secondary-topology fidelity by $52.7\%$ over the strongest baseline, while generating diverse non-native designs.

\section{Problem Setup}
\label{sec:problem_setup}

\paragraph{RNA sequence, topology, and geometry.}
An RNA sequence is a length-$L$ string $x=(x_1,\ldots,x_L)$ over the alphabet $\mathcal{V}=\{\texttt{A},\texttt{U},\texttt{G},\texttt{C}\}$.  We represent the target secondary topology as $\mathcal{T}^\star=(\mathcal{P}^\star,\mathcal{S}^\star)$, where $\mathcal{P}^\star\subseteq\{(p,q):1\le p<q\le L\}$ is the set of target base pairs and $\mathcal{S}^\star$ is the set of unpaired positions. The same topology can be written in dot-bracket notation; pseudoknots correspond to crossing pairs in $\mathcal{P}^\star$ and are represented with additional bracket types. Unless otherwise stated, $\mathcal{P}^\star$ denotes the canonical and wobble base-pairing scaffold of the target. Non-canonical tertiary contacts are not imposed as discrete pair actions, but are evaluated through the tertiary geometry objective.

The target tertiary structure is denoted by $\mathcal{G}^\star$. We use a $3$-bead coarse-grained representation~\cite{dawson2016coarse}, where each nucleotide is described by the Cartesian coordinates of the backbone atoms $\mathrm{P}$ and $\mathrm{C4'}$ and an unlabeled glycosidic nitrogen bead ($\mathrm{N1}$ for pyrimidines and $\mathrm{N9}$ for purines in the source structure). The model receives geometric coordinates but not native nucleotide identities. This representation compactly captures backbone conformation and approximate base placement for tertiary-structure-conditioned design.
 
\paragraph{Forward oracles.}
Given a candidate sequence $x$, a tertiary-structure oracle
$F_{\mathrm{3D}}$ predicts a fold $\widehat{\mathcal{G}}(x)=F_{\mathrm{3D}}(x)$. We distinguish two secondary-structure oracles. During topology-only initialization, we use a secondary folding oracle $\widehat{\mathcal{T}}^{\mathrm{fold}}(x)=F^{\mathrm{fold}}_{\mathrm{2D}}(x)$, which predicts secondary structure directly from sequence. During geometry-conditioned training and evaluation, we derive secondary topology from the predicted tertiary structure: $\widehat{\mathcal{T}}^{\mathrm{ann}}(x)=A_{\mathrm{2D}}\!\left(F_{\mathrm{3D}}(x)\right)$, where $A_{\mathrm{2D}}$ is a $\mathrm{3D}$-to-$\mathrm{2D}$ annotation procedure such as DSSR~\cite{lu2015dssr}. This distinction lets the first stage learn from a cheaper topology objective, while later stages measure topology and geometry under the same fold-back structure.

\paragraph{Topology-and-geometry co-design.}
RNA $\mathrm{3D}$ inverse design is commonly evaluated by tertiary fold-back fidelity. We instead formulate a co-design objective that requires both tertiary-structure fidelity and secondary-topology fidelity. For a completed sequence $x$, we define
\begin{equation}
    R_{\mathrm{3D}}(x;\mathcal{G}^\star)=\operatorname{TM}\!\left(\widehat{\mathcal{G}}(x),\mathcal{G}^\star\right),
    \label{eq:3D_eq}
\end{equation}
and
\begin{equation}
    R_{\mathrm{2D}}(x;\mathcal{T}^\star)=\operatorname{BP\text{-}F1}\!\left(\widehat{\mathcal{T}}^  {\mathrm{ann}}(x),\mathcal{T}^\star\right),
    \label{eq:2D_eq}    
\end{equation}
where $\mathrm{TM}$-score measures global tertiary alignment and $\mathrm{BP\text{-}F1}$ measures recovery of the target base pairs. A sequence is considered a co-design success when it satisfies both
\begin{equation}
    R_{\mathrm{3D}}(x;\mathcal{G}^\star)\ge \tau_{\mathrm{3D}},\qquad R_{\mathrm{2D}}(x;\mathcal{T}^\star)\ge \tau_{\mathrm{2D}}.
\end{equation}
Unlike prior sequence--structure co-design~\cite{ma2025riboflow}, which jointly generates sequences and structures, our formulation designs RNA sequences to recover prescribed topology and geometry targets. Beyond finding a single high-scoring sequence, we seek a diverse set of candidates that achieve favorable trade-offs among topology fidelity, geometry fidelity, and sequence novelty.

\section{Method}
\label{sec:method}

\begin{figure*}[t]
    \centering
    \includegraphics[width=\textwidth]{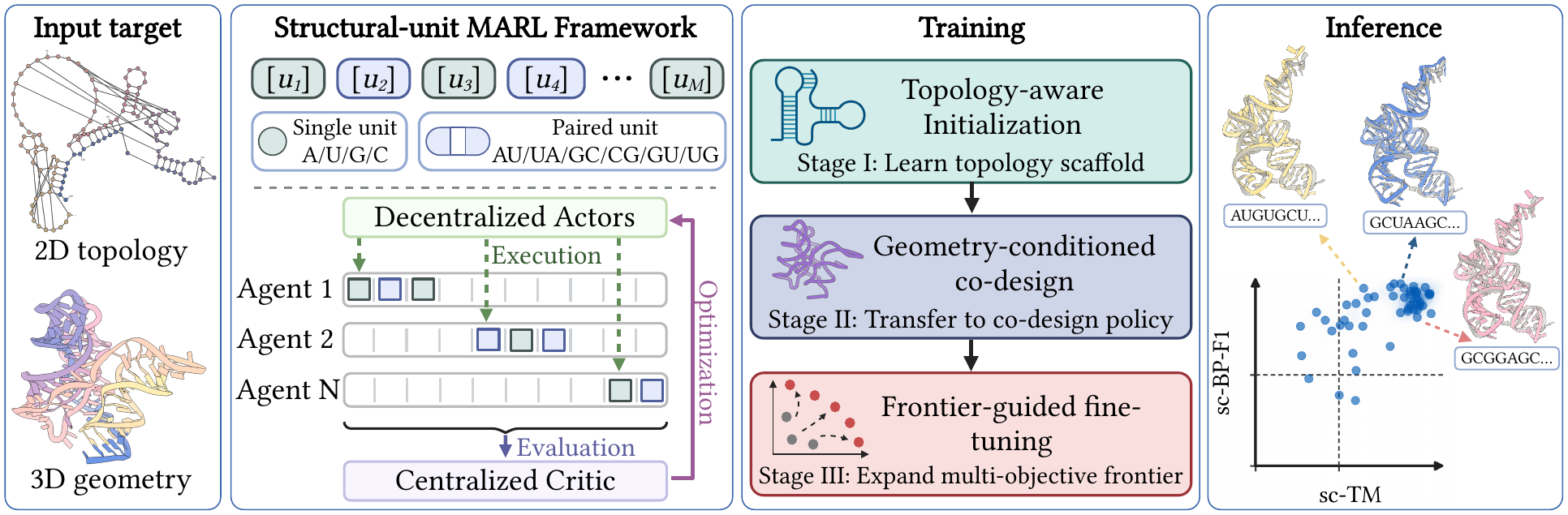}
    \caption{Overview of \our. \our\ formulates RNA design as cooperative sequence generation over structural units and follows a three-stage training pipeline.}
    \label{fig:pipeline}
\end{figure*}

\our\ is a cooperative reinforcement learning framework for RNA topology-and-geometry co-design. Given a target pair $(\mathcal{G}^\star,\mathcal{T}^\star)$, \our\ generates RNA sequences by filling topology-derived structural units rather than individual nucleotides.  Unpaired residues are designed as single-position units, while paired residues are designed jointly as paired units. This unit-level factorization exposes the base-pair couplings in the target topology and shortens the decision horizon.

Training proceeds in three stages. First, topology-aware initialization learns a policy for satisfying secondary-structure scaffolds. Second, geometry-conditioned co-design transfers this policy to tertiary-structure targets by adding $\mathrm{3D}$ geometric conditioning and optimizing a joint topology-and-geometry reward. Third, frontier-guided fine-tuning adapts the general policy to each target by improving the Pareto frontier over topology fidelity, geometry fidelity, and sequence diversity. \pref{fig:pipeline} illustrates the overall pipeline. Pseudocode is provided in~\pref{app:algorithms}.

\subsection{Structural-unit Cooperative Generation}
\label{sec:unit_generation}

\paragraph{Structural units.}
Given the target topology $\mathcal{T}^\star=(\mathcal{P}^\star,\mathcal{S}^\star)$, we construct an ordered list of structural units $\mathcal{U}^\star=(u_1,\ldots,u_M)$. Each unpaired position $p\in\mathcal{S}^\star$ forms a single-position unit $u=\{p\}$, and each target base pair $(p,q)\in\mathcal{P}^\star$ forms a paired unit $u=\{p,q\}$. Crossing pairs from pseudoknots are handled identically because the representation is set-based rather than nested. The units cover the full sequence and assign each residue to exactly one unit.

The action space depends on the unit type:
\begin{equation}
\mathcal{A}_{\mathrm{single}}=\{\texttt{A},\texttt{U},\texttt{G},\texttt{C}\},\qquad
\mathcal{A}_{\mathrm{pair}}=\{\texttt{AU},\texttt{UA},\texttt{GC},\texttt{CG},\texttt{GU},\texttt{UG}\}.
\end{equation}
Thus, paired units are filled by coordinated canonical or wobble base-pair assignments. This action factorization encodes the explicit secondary scaffold; non-canonical tertiary interactions are left to the $\mathrm{3D}$ structural reward.

\paragraph{Multi-agent generation.}
The ordered unit list is split into $N$ contiguous segments, one for each agent.  Agent $i$ maintains a cursor over its segment. At environment step $t$, every active agent observes its current unit, samples an action for that unit, writes the corresponding nucleotide assignment into the partial sequence, and advances its cursor. The episode terminates when all units have been assigned. Since up to $N$ units are filled per step, the effective horizon is approximately $\lceil M/N\rceil$ rather than $M$.

All agents share a single actor network with unit-type-specific output heads.  A centralized critic observes a pooled representation of the full partial design state during training, while each agent acts from its local observation at inference time. We optimize the policy under centralized training and decentralized execution. A formal MARL formulation and the training objective are provided in~\pref{app:rl_formulation}.

\paragraph{Topology-aware unit embeddings.}
For each residue position $p$, we embed the dot-bracket token and structural metadata $\boldsymbol r_p=E_{\mathrm{res}}(b_p,\boldsymbol m_p)$, where $\boldsymbol m_p$ includes pairing status, normalized position, pairing distance, and pseudoknot indicators. A single-position unit $u_j=\{p\}$ receives embedding
\begin{equation}
    \boldsymbol e_j=E_{\mathrm{single}}\!\left([\boldsymbol r_p,\boldsymbol m_{u_j}]\right),
\end{equation}
and a paired unit $u_j=\{p,q\}$ receives embedding
\begin{equation}
\boldsymbol e_j=E_{\mathrm{pair}}\!\left([\boldsymbol r_p,\boldsymbol r_q,\boldsymbol r_p\odot \boldsymbol r_q,|\boldsymbol r_p-\boldsymbol r_q|,\boldsymbol m_{u_j}]\right).
\end{equation}
The actor observation for agent $i$ at step $t$ is $o_t^i=[\boldsymbol e_{j_t^i},\boldsymbol q_{t,j_t^i},\boldsymbol c_{j_t^i}]$, where $j_t^i$ is the current unit index, $\boldsymbol q_{t,j_t^i}$ encodes the current assignment state of the unit, and $\boldsymbol c_{j_t^i}$ is a local context vector obtained by mean pooling embeddings in a fixed window around $u_{j_t^i}$.

\subsection{Stage I: Topology-aware Initialization}
\label{sec:topo_init}

The first stage trains the actor to construct sequences that satisfy target secondary topologies. After a complete sequence $x_T$ is generated, the topology-only oracle predicts $\widehat{\mathcal{T}}^{\mathrm{fold}}(x_T)=F^{\mathrm{fold}}_{\mathrm{2D}}(x_T)$. The terminal reward is
\begin{equation}
    R_{\mathrm{topo}}(x_T;\mathcal{T}^\star)=\operatorname{BP\text{-}F1}\left(\widehat{\mathcal{T}}^{\mathrm{fold}}(x_T),\mathcal{T}^\star\right)+\beta_{\mathrm{ex}}\mathbb{I}\left[\widehat{\mathcal{T}}^{\mathrm{fold}}(x_T)=\mathcal{T}^\star\right].
\end{equation}
All agents receive this shared reward. The policy is trained by maximizing
\begin{equation}
    \max_{\theta_{\mathrm{topo}}}\;\mathbb{E}_{x_T\sim \pi_{\theta_{\mathrm{topo}}}}\left[R_{\mathrm{topo}}(x_T;\mathcal{T}^\star)\right].
\end{equation}
This stage learns coordinated nucleotide assignments for single and paired structural units, providing a scaffolded initialization for subsequent $\mathrm{3D}$ co-design.

\subsection{Stage II: Geometry-conditioned Co-design}
\label{sec:geo_codesign}

The second stage transfers the topology-aware policy to tertiary-structure targets. The structural units and action spaces remain unchanged, but each unit is augmented with geometric information in $\mathcal{G}^\star$.

\paragraph{Geometry encoder.}
We encode the target tertiary structure with a \texttt{GVP-GNN}~\cite{jing2021learning} over the $3$-bead representation:
\begin{equation}
    \boldsymbol h^{\mathcal{G}}_1,\ldots,\boldsymbol h^{\mathcal{G}}_L=E_{\mathrm{GVP}}(\mathcal{G}^\star).
\end{equation}
Residue-level embeddings are pooled into unit-level geometric embeddings. For a single-position unit $u_j=\{p\}$, we set $\boldsymbol g_j=\boldsymbol h^{\mathcal{G}}_p$. For a paired unit
$u_j=\{p,q\}$, we use
\begin{equation}
    \boldsymbol g_j=\frac{1}{2}\left(\boldsymbol h^{\mathcal{G}}_p+\boldsymbol h^{\mathcal{G}}_q \right).
\end{equation}
A local $\mathrm{3D}$ context vector is computed by pooling spatial neighbors of each unit:
\begin{equation}
    \boldsymbol c^{\mathrm{3D}}_j=\operatorname{MeanPool}\left(\{\boldsymbol g_k: k\in \mathcal{W}_{\mathrm{3D}}(j)\}\right).
\end{equation}

\paragraph{Geometry-conditioned policy.}
For agent $i$ at unit $j_t^i$, the geometry-conditioned observation is $o_{t,\mathrm{geo}}^i=[o_t^i,\boldsymbol g_{j_t^i},\boldsymbol c^{\mathrm{3D}}_{j_t^i}]$. The actor is initialized from the topology-aware policy and augmented with two zero-initialized $\mathrm{3D}$-conditioned adapters. The final logits are
\begin{equation}
    \ell_i=\ell_i^{\mathrm{topo}}+\Delta \ell_i^{\mathrm{FiLM}}+\Delta \ell_i^{\mathrm{3D}},
\end{equation}
where $\ell_i^{\mathrm{topo}}$ is the inherited topology-aware logit vector, $\Delta \ell_i^{\mathrm{FiLM}}$ is produced by FiLM modulation of the hidden state using $\boldsymbol g_{j_t^i}$, and $\Delta \ell_i^{\mathrm{3D}}$ is a direct geometric correction. Zero initialization makes the transferred policy start near the topology-aware policy, so $\mathrm{3D}$ conditioning initially acts as a small residual adjustment rather than overwriting the learned topology scaffold.

\paragraph{Co-design reward.}
At the end of an episode, the completed sequence $x_T$ is folded by the $\mathrm{3D}$ oracle: $\widehat{\mathcal{G}}(x_T)=F_{\mathrm{3D}}(x_T)$ and $\widehat{\mathcal{T}}^{\mathrm{ann}}(x_T)=A_{\mathrm{2D}}\!\left(\widehat{\mathcal{G}}(x_T)\right)$. Based on Eqs.~\eqref{eq:3D_eq} and~\eqref{eq:2D_eq}, we compute the Stage-II terminal reward as
\begin{equation}
    R_{\mathrm{co}}(x_T)=R_{\mathrm{3D}}(x_T;\mathcal{G}^\star)+\lambda_{\mathrm{2D}}R_{\mathrm{2D}}(x_T;\mathcal{T}^\star)-\lambda_{\mathrm{viol}}\left[\tau_{\mathrm{2D}}-R_{\mathrm{2D}}(x_T;\mathcal{T}^\star)\right]_+.
\end{equation}
The positive $R_{\mathrm{2D}}$ term encourages topology recovery, while the hinge penalty discourages geometry-improving updates that fall below the desired topology-fidelity threshold. This stage produces a general topology-and-geometry co-design policy.

\subsection{Stage III: Frontier-guided Fine-tuning}
\label{sec:frontier}

The final stage performs target-wise policy adaptation. This shifts costly trial-and-error exploration from the wet lab to \emph{in silico} optimization, prioritizing promising candidates for subsequent experimental validation. The trade-offs among topology fidelity, geometric fidelity, and sequence diversity motivate a multi-objective reinforcement learning formulation~\citep{roijers2013survey,hu2023momix,yang2019generalized,hu2024pa2d,hayes2022practical,hu2026human}, rather than optimization under a single fixed scalarization. \our\ therefore fine-tunes the policy to expand the target-specific Pareto frontier.

\paragraph{Archive and objective vectors.}
For each target $(\mathcal{G}^\star,\mathcal{T}^\star)$, we maintain an archive of evaluated sequences, $\mathcal{H}_k^\star=\{x^{(1)},\ldots,x^{(K)}\}$, at fine-tuning iteration $k$. The archive is warm-started with candidates sampled from the Stage-II co-design policy. Each sequence is mapped to an objective vector. The default bi-objective setting uses
\begin{equation}
    \boldsymbol r(x)=[R_{\mathrm{2D}}(x;\mathcal{T}^\star),R_{\mathrm{3D}}(x;\mathcal{G}^\star)].
\end{equation}
We also consider a diversity-aware tri-objective setting:
\begin{equation}
    \boldsymbol r_{\mathrm{div}}(x)=[R_{\mathrm{2D}}(x;\mathcal{T}^\star),R_{\mathrm{3D}}(x;\mathcal{G}^\star),R_{\mathrm{div}}(x;\mathcal{H}_k^\star)],
\end{equation}
where
\begin{equation}
    R_{\mathrm{div}}(x;\mathcal{H}_k^\star)=\min_{x'\in\mathcal{H}_k^\star}\frac{1}{L}\sum_{p=1}^{L}\mathbb{I}[x_p\neq x'_p].
\end{equation}
All objectives are normalized to $[0,1]$ with larger values preferred. Duplicate sequences receive zero diversity and are inserted into the archive only once.

A sequence $x$ dominates $y$ if it is no worse in every selected objective and strictly better in at least one. The non-dominated archive is $\mathcal{F}_k^\star=\operatorname{ND}(\mathcal{H}_k^\star)$.

\paragraph{Batch-level hypervolume reward.}
At iteration $k$, the current policy samples a batch of candidates $\mathcal{C}_k$. We evaluate their objective vectors and compute the post-batch frontier $\mathcal{P}_k=\operatorname{ND}(\mathcal{H}_k^\star\cup\mathcal{C}_k)$. Let $\mathcal{R}(\mathcal{A})=\{\boldsymbol r(x):x\in\mathcal{A}\}$ denote the set of objective vectors for a sequence set $\mathcal{A}$. With reference point $\boldsymbol r_{\mathrm{ref}}$ in the selected objective space, the frontier reward for candidate $x\in\mathcal{C}_k$ is its exclusive hypervolume contribution:
\begin{equation}
    R_{\mathrm{front}}(x)=\operatorname{HV}\left(\mathcal{R}(\mathcal{P}_k);\boldsymbol r_{\mathrm{ref}}\right)-\operatorname{HV}\left(\mathcal{R}\left(\operatorname{ND}\left(\mathcal{H}_k^\star\cup\left(\mathcal{C}_k\setminus\{x\}\right)\right)\right);\boldsymbol r_{\mathrm{ref}}\right).
\end{equation}
The batch-level definition makes the reward independent of the order in which candidates are inserted into the archive. A candidate receives positive reward only if it expands the target-wise frontier. After reward computation, all evaluated candidates are added to the archive to form $\mathcal{H}_{k+1}^\star$.

\paragraph{Fine-tuning objective.}
Starting from the Stage-II co-design policy, we optimize
\begin{equation}
    \max_{\theta_{\mathrm{front}}}\;\mathbb{E}_{x_T\sim \pi_{\theta_{\mathrm{front}}}}\left[R_{\mathrm{front}}(x_T)\right].
\end{equation}
The reward is terminal and shared across agents. The resulting policy is target-adapted: it is rewarded not for improving a fixed scalarized score, but for producing sequences that expand the achievable frontier of topology fidelity, geometry fidelity, and diversity.

\section{Experimental Setup}
\label{sec:experiment_setup}

\paragraph{Benchmark construction.}
We construct a topology-and-geometry co-design benchmark from experimentally resolved RNA structures in the PDB~\cite{pdb2000}. We retain single-chain RNA structures with resolution better than $4$~\AA, remove structures with chain breaks, and deduplicate targets by secondary topology. To make predictor-based fold-back evaluation meaningful, we refold each native sequence with both \texttt{RhoFold}~\citep{shen2024accurate} and \texttt{AlphaFold3}~\citep{abramson2024accurate}, and retain only targets whose native sequence recovers the experimental fold with $\mathrm{TM\text{-}score} \ge 0.45$ under both predictors, a threshold commonly used to indicate shared global RNA topology~\citep{zhang2022us}. This filtering yields $1{,}213$ targets. It reduces ambiguity in fold-back evaluation, but also biases the benchmark toward RNAs that current structure predictors can fold reasonably well.

We split targets by structural dissimilarity rather than random sequence-level partitioning. Specifically, we cluster targets by pairwise tertiary-structure similarity using $k$-medoids and choose held-out targets with the lowest nearest-neighbor similarity to the training set. The final test set contains $50$ structurally remote targets, with mean nearest-neighbor similarity $0.34$ and maximum nearest-neighbor similarity $0.42$ to the training set. Among these targets, $44\%$ contain pseudoknotted base pairs.

\paragraph{Evaluation protocol and metrics.}
We instantiate the co-design evaluation protocol defined in Section~\ref{sec:problem_setup}. Unless otherwise stated, main results use \texttt{RhoFold} as the fold-back oracle; cross-oracle experiments evaluate the same designed sequences with \texttt{AlphaFold3}. For each target, we first fold the native sequence with the selected oracle to obtain a predictor-specific reference structure. We then fold each designed sequence with the same oracle and compare the two predicted structures. This self-consistency protocol avoids mixing experimental structures and predictor-specific folds in the primary metric.

We report tertiary fidelity by self-consistency $\mathrm{TM\text{-}score}$ ($\mathrm{sc\text{-}TM}$) and self-consistency $\mathrm{RMSD}$ ($\mathrm{sc\text{-}RMSD}$). We report secondary-topology fidelity by extracting base pairs from the predicted designed structure with DSSR~\citep{lu2015dssr} and computing self-consistency base-pair F1 ($\mathrm{sc\text{-}BP\text{-}F1}$) against the target topology. Our primary task-level metric is co-design success: a candidate is successful only if it satisfies both
\begin{equation}
\mathrm{sc\text{-}TM}> \tau_{\mathrm{3D}}
\quad \mathrm{and} \quad
\mathrm{sc\text{-}BP\text{-}F1}> \tau_{\mathrm{2D}}. 
\end{equation}
We use $\tau_{\mathrm{3D}}=0.45$ and $\tau_{\mathrm{2D}}=0.50$ in the main experiments, and report threshold sensitivity in Appendix~\ref{app:threshold_sensitivity_results}. The co-design solved target rate is the fraction of test targets for which at least one generated candidate satisfies both thresholds.

\paragraph{Baselines and inference budget.}
We compare against recent $\mathrm{3D}$ RNA inverse-design methods: \texttt{gRNAde}~\cite{joshi2025grnade}, \texttt{RDesign}~\cite{tan2024rdesign}, \texttt{R3Design}~\cite{tan2025r3design}, \texttt{RiboDiffusion}~\cite{huang2024ribodiffusion}, and \texttt{RIDER}~\cite{hu2026rider}. All baselines are retrained on the same training split and evaluated on the same held-out targets. Each method generates $K=10$ candidate sequences per target under its default sampling procedure.

\paragraph{Metrics and statistics.}
For continuous metrics, we first aggregate the $K$ candidates for each target using mean-of-$K$, and then report mean $\pm$ standard deviation across targets. For success metrics, we report candidate-level success rate and target-level solved rate separately. For $\mathrm{sc}\text{-}\mathrm{TM}$, $\mathrm{sc}\text{-}\mathrm{BP}\text{-}\mathrm{F1}$, and success rate, we compute $p$-values using a two-sided paired Wilcoxon signed-rank test over target-level scores, followed by Holm correction for multiple comparisons. For diversity metrics and success-conditioned novelty, where the set of valid targets can differ across methods, we use a two-sided Wilcoxon rank-sum test. For solved target rates, we use a two-proportion $z$-test.
\section{Results}
\label{sec:results}

\subsection{\our\ improves topology and geometry fidelity}

We first evaluate whether \our\ improves RNA $\mathrm{3D}$ inverse design under the co-design criterion defined in~\pref{sec:experiment_setup}. \pref{tab:main} compares \our\ with five recent $\mathrm{3D}$ RNA inverse-design methods: \texttt{gRNAde}, \texttt{RDesign}, \texttt{R3Design}, \texttt{RiboDiffusion}, and \texttt{RIDER}. All methods are trained on the same split and evaluated on the same structurally remote test targets.

\begin{table}[t!]
\centering
\renewcommand{\arraystretch}{1.1}
\small
\caption{Structural design performance on the test set. Each method generates $K=10$ candidates per target. Continuous metrics report mean $\pm$ standard deviation over target-level aggregated scores. Success denotes \emph{co-design success}, defined as $\mathrm{sc\text{-}TM}>0.45$ and $\mathrm{sc\text{-}BP\text{-}F1}>0.50$. Solved targets denotes the fraction of targets for which at least one candidate satisfies both thresholds. Threshold sensitivity is reported in~\pref{app:threshold_sensitivity_results}.}
\label{tab:main}
\resizebox{\textwidth}{!}{
\begin{tabular}{@{}lccccc@{}}
\toprule
\textbf{Method} & \textbf{$\mathrm{sc\text{-}TM}$} $\uparrow$ & \textbf{$\mathrm{sc\text{-}BP\text{-}F1}$} $\uparrow$ & \textbf{$\mathrm{sc\text{-}RMSD}$} $\downarrow$ & \textbf{Success rate} $\uparrow$ & \textbf{Solved targets} $\uparrow$ \\
\midrule
\texttt{gRNAde} & $0.32 \pm 0.12$ & $0.36 \pm 0.26$ & $3.54 \pm 1.00$ & $0.17$ & $38.00\%$ \\
\texttt{RDesign} & $0.31 \pm 0.10$ & $0.31 \pm 0.21$ & $3.61 \pm 0.91$ & $0.13$ & $42.00\%$ \\
\texttt{R3Design} & $0.35 \pm 0.14$ & $0.33 \pm 0.27$ & $3.47 \pm 1.03$ & $0.21$ & $50.00\%$ \\
\texttt{RiboDiffusion} & $0.32 \pm 0.13$ & $0.30 \pm 0.25$ & $3.58 \pm 1.14$ & $0.17$ & $42.00\%$ \\
\texttt{RIDER} & $0.60 \pm 0.13$ & $0.55 \pm 0.22$ & $2.94 \pm 1.26$ & $0.49$ & $60.00\%$ \\
\midrule
\our & $\mathbf{0.71 \pm 0.17}$ & $\mathbf{0.84 \pm 0.15}$ & $\mathbf{2.16 \pm 1.07}$ & $\mathbf{0.86}$ & $\mathbf{88.00\%}$ \\
\bottomrule
\end{tabular}
}
\end{table}

\begin{figure}[t!]
\centering
\includegraphics[width=1.0\textwidth]{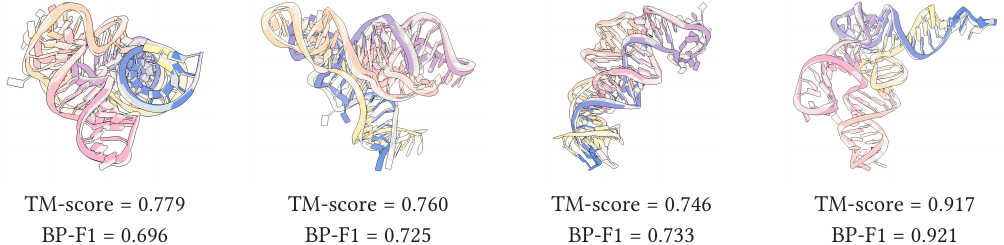}
\caption{Representative co-design successes for PDB targets \texttt{3IQP}, \texttt{6WJS}, \texttt{3UCU}, and \texttt{3ADB}. Gray denotes the target reference structure, and color denotes the fold-back structure of the designed sequence. All shown designs satisfy the joint co-design success criterion.}
\label{fig:visualization}
\end{figure}

\our\ achieves the strongest performance on both topology and geometry metrics. Relative to the strongest baseline, \texttt{RIDER}, \our\ improves $\mathrm{sc\text{-}TM}$ from $0.60$ to $0.71$ and $\mathrm{sc\text{-}BP\text{-}F1}$ from $0.55$ to $0.84$, corresponding to relative gains of $18.3\%$ and $52.7\%$, respectively. The improvement in $\mathrm{sc\text{-}BP\text{-}F1}$ is central to our setting: the gain in tertiary fold-back fidelity is accompanied by substantially better recovery of the target base-pairing topology. \our\ also obtains the lowest $\mathrm{sc\text{-}RMSD}$, indicating lower structural deviation after alignment.

The joint success metrics provide the primary evidence for topology-and-geometry co-design. \our\ improves the co-design success rate from $0.49$ to $0.86$ and the solved target rate from $60.00\%$ to $88.00\%$ compared with \texttt{RIDER}. Because a successful candidate must satisfy both the geometry and topology thresholds, these results support our formulation of RNA $\mathrm{3D}$ inverse design as a joint co-design problem rather than a geometry-only fold-back task.  Representative co-design successes are shown in~\pref{fig:visualization}.

\subsection{Target-wise frontier-guided refinement expands the co-design frontier}

\begin{figure}[t!]
\centering
\includegraphics[width=\linewidth]{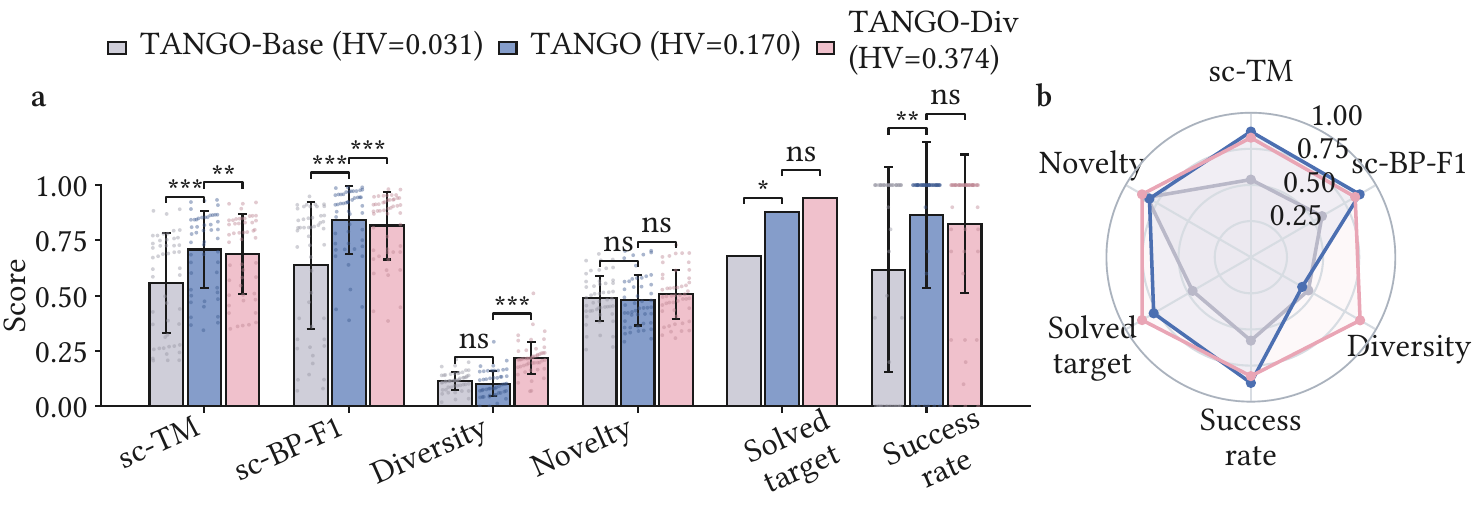}
\caption{Target-wise frontier-guided refinement. (a) Structural quality and sequence diversity for the stage-II policy (\texttt{TANGO-Base}), the bi-objective frontier-refined model (\our{}), and the diversity-aware tri-objective variant (\texttt{TANGO-Div}). (b) Multi-objective trade-off visualization between design quality and diversity. HV denotes the dominated hypervolume of each method's Pareto frontier in the normalized objective space. Significance markers denote \texttt{*} for $p<0.05$, \texttt{**} for $p<0.01$, \texttt{***} for $p<0.001$, and ns for not significant. }
\label{fig:composite_1}
\end{figure}

We next isolate the contribution of the target-wise refinement stage. \pref{fig:composite_1}(a) compares the stage-II policy before target-wise fine-tuning (\texttt{TANGO-Base}), the default bi-objective frontier-refined model (\our), and the diversity-aware tri-objective variant (\texttt{TANGO-Div}). Compared with \texttt{TANGO-Base}, \our\ improves $\mathrm{sc\text{-}TM}$ by $27.6\%$ and $\mathrm{sc\text{-}BP\text{-}F1}$ by $31.9\%$, indicating that the target-wise refinement stage contributes to both tertiary fold-back fidelity and secondary-topology recovery. The improvements in success rate and solved target rate further show that the refined policies place more candidates in the high-\(\mathrm{sc\text{-}BP\text{-}F1}\), high-\(\mathrm{sc\text{-}TM}\) region, rather than improving only one objective. 

Adding sequence diversity as an explicit objective increases diversity by $90.1\%$ relative to \texttt{TANGO-Base}, while preserving competitive topology and geometry fidelity. \pref{fig:composite_1}(b) summarizes the resulting multi-objective trade-offs. The hypervolume increases from $0.031$ for \texttt{TANGO-Base} to $0.170$ for \our{} and $0.374$ for \texttt{TANGO-Div}, indicating that frontier-guided refinement expands the set of useful target-specific candidates.

To test whether this gain is merely due to additional test-time sampling, \pref{app:top10_sampling_results} compares against sampling-based re-ranking under the same candidate budget, and~\pref{app:weighted_reward_results} compares against fixed weighted-reward fine-tuning. These controls suggest that the improvement comes from optimizing the target-wise frontier, rather than only from selecting better samples from an unchanged policy or using a fixed scalarization of the objectives.

\subsection{Topology-aware initialization provides a useful scaffold for \texorpdfstring{$\mathrm{3D}$}{3D} design}
\begin{figure}[!tbp]
  \centering
  \includegraphics[width=0.50\linewidth]{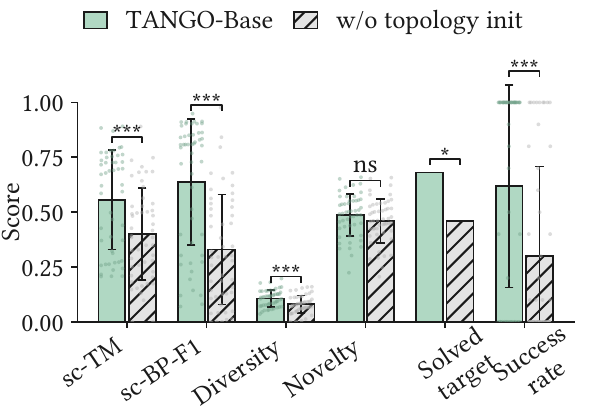}
  \caption{Effect of topology-aware initialization. The ablated variant removes Stage-I topology pretraining and trains the geometry-conditioned policy from scratch, while keeping the structural-unit action space and model architecture unchanged.}
  \label{fig:topology_init_ablation}
\end{figure}

Here we ablate the Stage-I topology-aware initialization by using the same structural-unit action space and geometry-conditioned architecture, but starting $\mathrm{3D}$ co-design training without first learning to satisfy secondary-structure scaffolds. \pref{fig:topology_init_ablation} shows that removing this initialization substantially degrades all primary co-design metrics. With topology-aware initialization, \texttt{TANGO-Base} achieves relative improvements of $38.9\%$ in $\mathrm{sc\text{-}TM}$, $93.3\%$ in $\mathrm{sc\text{-}BP\text{-}F1}$, and $106.0\%$ in co-design success rate.

This ablation supports the curriculum design. Although Stage I optimizes only a secondary-topology objective, it improves both topology recovery and tertiary fold-back fidelity after $\mathrm{3D}$ training.

This is consistent with the intuition that a topology-compatible initialization places the policy in a better exploration region for subsequent tertiary optimization, where fold-back rewards are less sparse and base-paired positions are already coordinated.

\subsection{\our\ generates non-native high-quality designs}

\begin{figure}[!tbp]
\centering
\includegraphics[width=1.0\linewidth]{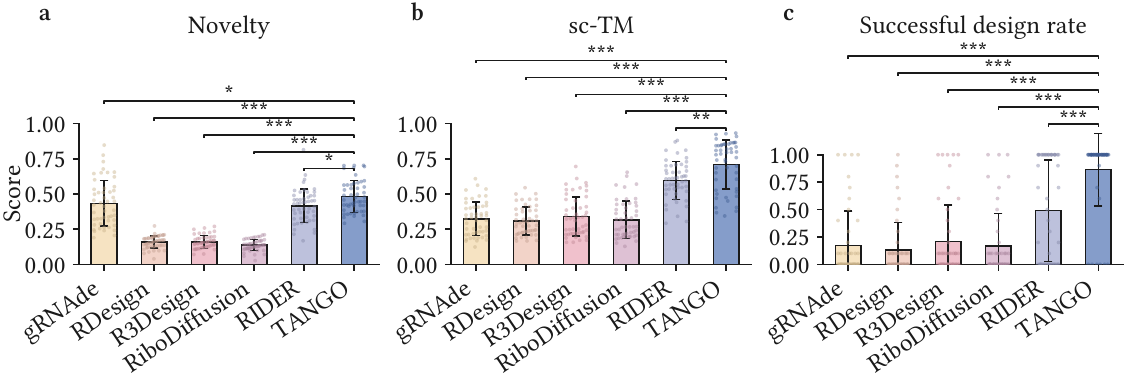}
\caption{Sequence novelty and design quality across methods. Novelty is measured as normalized Hamming distance to the native target sequence. Design quality is reported by $\mathrm{sc\text{-}TM}$ and co-design success rate, where success requires both topology and geometry thresholds.}
\label{fig:composite_2}
\end{figure}

We next examine whether high fold-back scores are achieved by recovering native-like sequences. \pref{fig:composite_2}(a) reports sequence novelty, defined as the normalized Hamming distance between a designed sequence and the native sequence of the target. \pref{fig:composite_2}(b) and~\pref{fig:composite_2}(c) report the corresponding $\mathrm{sc\text{-}TM}$ and co-design success rate, respectively. \our\ achieves the strongest combination of novelty and design quality among the compared methods: its generated sequences remain far from the native target sequences while maintaining the highest co-design success rate.

This result suggests that \our\ can identify non-native sequence solutions that preserve both target topology and tertiary geometry. However, distance-to-native alone does not rule out all forms of memorization or predictor-specific sequence bias; we therefore interpret this analysis as evidence of non-native design rather than as definitive proof of memorization-free generation.

\subsection{Robustness across oracles and architectures}
\label{sec:robustness_results}

We finally stress-test whether the main conclusions depend on a single fold-back oracle or architecture choice.

\begin{figure}[!tbp]
  \centering
  \includegraphics[width=0.40\linewidth]{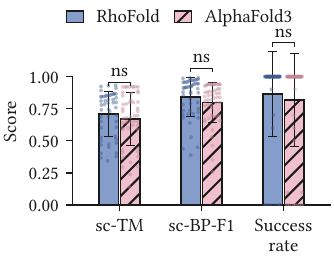}
  \caption{Cross-oracle evaluation of the same \our\ designs using \texttt{RhoFold} and \texttt{AlphaFold3}. No \texttt{AlphaFold3}-based refinement is performed.}
  \label{fig:cross_oracle}
\end{figure}

First, \pref{fig:cross_oracle} evaluates the same \our-generated sequences under both \texttt{RhoFold} and \texttt{AlphaFold3}. The $\mathrm{sc\text{-}TM}$ and $\mathrm{sc\text{-}BP\text{-}F1}$ scores remain comparable across the two predictors, suggesting that \our{} designs are not obvious artifacts of one fold-back oracle. This remains predictor-based validation rather than experimental folding validation.

Second, \pref{app:multi_agent_results} ablates the multi-agent architecture. Compared with a single-agent variant, the multi-agent policy improves both topology and geometry metrics, supporting the use of topology-derived structural units and parallel cooperative generation for RNA co-design.

\section{Conclusion}
\label{sec:conclusion}

This work argues that RNA $\mathrm{3D}$ inverse design should not be treated as a geometry-only fold-back problem. Because RNA tertiary structure is scaffolded by base-pairing topology, useful designs should recover both the target $\mathrm{3D}$ fold and the target secondary topology. We operationalize this view through \our, a curriculum MARL framework that designs topology-derived structural units, transfers a secondary-structure policy to geometry-conditioned co-design, and uses frontier-guided fine-tuning to improve target-wise trade-offs among topology fidelity, geometry fidelity, and sequence diversity. Experiments demonstrated that \our\ improves fold-back geometry, topology recovery, and joint co-design success over recent $\mathrm{3D}$ RNA design baselines. Future work could extend \our\ to more complex biological settings and experimentally validate selected designs.

\section*{Acknowledgments}

This work was supported by the UKRI Future Leaders Fellowship under grants MR/S017062/1 and MR/X011135/1; the National Natural Science Foundation of China under grants 62376056, 62076056 and 62625312; the Science and Technology Innovation Program of Hunan Province under Grant 2024RC1011; the Major Basic Research Projects in Hunan Province under Grant 2026JC0001; the Isambard AI Open Access (AIOA - 01 - 2026-06-04 - EvoRNA: Predictive Scaling of Evolution-Aware and Multimodal RNA Foundation Models for Crop Resilience); the Modal compute grant; the Royal Society Faraday Discovery Fellowship (FDF/S2/251014); BBSRC Transformative Research Technologies (UKRI1875); the Royal Society International Exchanges Award (IES/R3/243136); the Kan Tong Po Fellowship (KTP/R1/231017); the Amazon Research Award; and the Alan Turing Fellowship.

\clearpage
\bibliographystyle{unsrtnat}
\bibliography{references/refs}

\appendix
\setcounter{figure}{0}
\renewcommand{\thefigure}{A\arabic{figure}}
\renewcommand{\theHfigure}{appendix.\arabic{figure}}
\setcounter{table}{0}
\renewcommand{\thetable}{A\arabic{table}}
\renewcommand{\theHtable}{appendix.\arabic{table}}
\newpage
\begingroup

\raggedbottom
\renewcommand{\topfraction}{0.95}
\renewcommand{\textfraction}{0.05}
\renewcommand{\floatpagefraction}{0.9}
\setcounter{topnumber}{4}
\setcounter{totalnumber}{6}
\makeatletter
\setlength{\@fptop}{0pt}
\setlength{\@fpsep}{12pt plus 2pt minus 2pt}
\makeatother

\section{Related Work and Positioning}
\label{app:related_work}

\paragraph{RNA inverse design.}
Computational RNA inverse design seeks sequences that satisfy specified structural objectives. Early work primarily studied secondary-structure inverse folding, where the target is a dot-bracket topology. Local-search methods~\citep{HofackerFSBTS94,AndronescuFHHC04,BuschB06} iteratively modify sequences using folding feedback, metaheuristic approaches such as ant-colony optimization~\citep{kleinkauf2015antarna} and evolutionary algorithms~\citep{Taneda12,TaheriG15,Merleau2022,zhou2023rna} explore large combinatorial spaces, and reinforcement-learning methods~\citep{eastman2018solving,runge2019learning,runge2024partial} learn sequence-construction policies. \texttt{CocoRNA}~\citep{hu2026cocorna} applies cooperative multi-agent reinforcement learning to secondary-structure inverse design by distributing sequence construction among agents trained with a centralized critic. Many widely used secondary-design methods are built for pseudoknot-free structures, although methods such as \texttt{antaRNA}~\citep{kleinkauf2015antarna} and \texttt{aRNAque}~\citep{Merleau2022} support broader topology classes.  These methods directly optimize secondary topology, but do not by themselves enforce the tertiary geometry required by many functional RNA folds.

Recent work has shifted toward tertiary-structure-based RNA design. Rosetta provides physics-based fixed-backbone RNA design through energy modeling~\citep{leman2020macromolecular}. Deep learning methods further condition sequence generation on RNA $\mathrm{3D}$ structure: graph neural networks~\citep{tan2024rdesign,joshi2025grnade,tan2025r3design} encode geometric structure, diffusion models such as \texttt{RiboDiffusion}~\citep{huang2024ribodiffusion} learn conditional sequence distributions over $\mathrm{3D}$ backbones, and Bayesian flow networks~\citep{wang2025rbfn} provide another generative formulation. \texttt{RIDER}~\citep{hu2026rider} is closest to our use of structural optimization, as it fine-tunes a generative model with reinforcement learning to improve $\mathrm{3D}$ fold-back similarity.

Our work is complementary to these approaches. Existing $\mathrm{3D}$ RNA design methods primarily condition on tertiary geometry and are often evaluated by native sequence recovery or tertiary fold-back fidelity. Some methods incorporate secondary-structure information as priors, constraints, or repair mechanisms; for example, \texttt{RDesign} uses extracted secondary structures with base-pair information to guide design~\citep{tan2024rdesign}. However, topology fidelity has not been systematically formulated, optimized, and evaluated as a first-class objective jointly with tertiary-structure fidelity. \our\ addresses this gap by using the target topology to define structural-unit actions, initializing the policy on secondary-structure design, and refining candidates with multi-objective topology-and-geometry rewards.

\paragraph{RNA structure prediction as design oracles.}
RNA structure prediction is the forward problem complementary to inverse design.  Secondary-structure predictors include thermodynamic models such as \texttt{ViennaRNA}~\citep{lorenz2011viennarna} and \texttt{RNAstructure}~\citep{bellaousov2013rnastructure}, as well as deep learning models~\citep{singh2019spotrna,fu2022ufold,he2024ribonanza}. RNA foundation models~\citep{yang2025bridging,wang2024rnaernie,penic2025rinalmo, chen2022rnafm} provide pretrained representations for structure-related tasks~\citep{li2026transfer}. Tertiary-structure prediction has advanced from fragment assembly methods such as \texttt{RNAComposer}~\citep{popenda2012automated} to deep geometric predictors including \texttt{trRosettaRNA}~\citep{wang2023trrosettarna}, \texttt{DRFold}~\citep{li2023drfold}, \texttt{RhoFold}~\citep{shen2024accurate}, and \texttt{AlphaFold3}~\citep{abramson2024accurate}. Related generative models such as \texttt{RNAbpFlow}~\citep{tarafder2025rnabpflow} incorporate base-pairing conditions for RNA $\mathrm{3D}$ structure generation. \texttt{REFLEX}~\citep{ma2026reflex} generates sequence-conditioned RNA conformational ensembles using a flexibility-calibrated stochastic bridge.

In this work, structure predictors are used as fold-back oracles for evaluating designed sequences, not as the proposed design model itself. Because predictor-based evaluation can introduce oracle bias, our experiments report cross-oracle validation with both \texttt{RhoFold} and \texttt{AlphaFold3}.

\section{Algorithm Pseudocode}
\label{app:algorithms}

This section provides pseudocode for the main algorithmic components of \our. We include three procedures:~\pref{alg:tango_units} constructs structural units from the target topology,~\pref{alg:tango_rollout} performs multi-agent sequence generation over these units, and~\pref{alg:tango_training} summarizes the shared MARL training framework used across topology-aware initialization, topology-and-geometry co-design, and frontier-guided fine-tuning.

\subsection{Structural-unit Cooperative Generation}
\label{app:unit_algorithm}

TANGO first converts the target topology into an ordered list of structural units, and then assigns these units to multiple agents for cooperative sequence generation.~\pref{alg:tango_units} gives the pseudocode for constructing structural units from a target pair map.

\begin{algorithm}
\caption{\our: Constructing Structural Units}
\label{alg:tango_units}
\begin{algorithmic}[1]
\tangoalgspacing
\REQUIRE Pair map $m^\star$ for target topology $\mathcal{T}^\star$, with $m^\star_p=q$ for paired residues and $m^\star_p=0$ for unpaired residues
\ENSURE Ordered structural units $\mathcal{U}^\star=(u_1,\ldots,u_M)$
\STATE Initialize empty unit list $\mathcal{U}^\star$ and empty assigned set $\mathcal{V}$
\FOR{$p=1$ \textbf{to} $L$}
    \IF{$p\in\mathcal{V}$}
        \STATE Continue
    \ENDIF
    \IF{$m^\star_p=q$ and $q>0$}
        \IF{$p<q$}
            \STATE Create paired unit $u=(p,q)$
            \STATE Store paired-unit metadata
            \STATE Append $u$ to $\mathcal{U}^\star$ and add $p,q$ to $\mathcal{V}$
        \ENDIF
    \ELSE
        \STATE Create single-position unit $u=(p)$
        \STATE Store single-unit metadata
        \STATE Append $u$ to $\mathcal{U}^\star$ and add $p$ to $\mathcal{V}$
    \ENDIF
\ENDFOR
\STATE \RETURN $\mathcal{U}^\star$
\end{algorithmic}
\end{algorithm}

The resulting unit list is used as the action schedule for multi-agent rollout. As shown in~\pref{alg:tango_rollout}, each agent is responsible for generating the sequence for its assigned units, and the joint action at each step consists of the nucleotide choices made by all active agents. The rollout proceeds until all units are assigned, at which point a complete sequence is returned for reward evaluation.

\begin{algorithm}
\caption{\our: Multi-Agent Structural-Unit Rollout}
\label{alg:tango_rollout}
\begin{algorithmic}[1]
\tangoalgspacing
\REQUIRE Structural units $\mathcal{U}^\star$, actor $\pi_\theta$, number of agents $N$
\ENSURE Completed sequence $x_T$ and rollout trajectory $\tau$
\STATE Split $\mathcal{U}^\star$ into $N$ contiguous agent segments
\STATE Initialize partial sequence $x_0$, assignment mask, and one cursor per agent
\STATE Initialize empty trajectory buffer $\tau$
\FOR{$t=0$ \textbf{to} $T-1$}
    \STATE Initialize empty joint action map $\boldsymbol{a}_t$
    \FOR{each agent $i=1,\ldots,N$}
        \IF{agent $i$ has no remaining unit}
            \STATE Mark agent $i$ inactive
        \ELSE
            \STATE Build local observation $o_t^i$ for the unit at agent $i$'s cursor
            \IF{$\mathrm{3D}$ conditions are available}
                \STATE Augment $o_t^i$ with unit-level $\mathrm{3D}$ information
            \ENDIF
            \STATE Sample $a_t^i\sim\pi_\theta(\cdot\mid o_t^i)$ using the corresponding unit-type head
            \STATE Store $a_t^i$ in $\boldsymbol{a}_t$ and record rollout information in $\tau$
        \ENDIF
    \ENDFOR
    \STATE Write active unit assignments from $\boldsymbol{a}_t$ into the partial sequence
    \STATE Advance cursors for active agents and update the assignment mask
    \IF{all units are assigned}
        \STATE Break
    \ENDIF
\ENDFOR
\STATE \RETURN completed sequence $x_T$ and trajectory $\tau$
\end{algorithmic}
\end{algorithm}

\subsection{Multi-Agent Reinforcement Learning Framework}
\label{app:training_algorithm}

Let $s\in\{\mathrm{I},\mathrm{II},\mathrm{III}\}$ denote the training stage. We use $\mathcal{R}_s$ to denote the stage-specific terminal reward: the topology-training reward for Stage I, the topology-and-geometry co-design reward for Stage II, and the frontier-guided reward for Stage III. All stages share the same structural-unit rollout interface and MAPPO update, while differing in the conditioning information and terminal reward construction. \pref{alg:tango_training} summarizes the overall training framework for \our.

\begin{algorithm}
\caption{\our: Multi-Agent Reinforcement Learning Framework}
\label{alg:tango_training}
\begin{algorithmic}[1]
\tangoalgspacing
\REQUIRE Training stage $s$, actor $\pi_\theta$, centralized critic $V_\psi$, MAPPO hyperparameters
\REQUIRE Training targets $\mathcal{D}$, structure oracles when required, and stage-specific reward $\mathcal{R}_s$
\ENSURE Updated policy parameters $\theta$
\FOR{each training update}
    \STATE Initialize rollout buffer $\mathcal{B}$
    \FOR{each sampled target in $\mathcal{D}$}
        \STATE Obtain target topology $\mathcal{T}^\star$ and construct $\mathcal{U}^\star$ using Algorithm~\ref{alg:tango_units}
        \STATE Build stage-specific conditioning information, including $\mathrm{3D}$ conditions when required
        \STATE Generate $(x_T,\tau)$ using Algorithm~\ref{alg:tango_rollout}
        \STATE Compute terminal reward $r_T=\mathcal{R}_s(x_T;\mathcal{G}^\star,\mathcal{T}^\star)$
        \STATE Assign shared terminal reward $r_T$ to active agent transitions in $\tau$
        \STATE Store $\tau$ in $\mathcal{B}$
    \ENDFOR
    \STATE Compute returns and GAE advantages using centralized critic $V_\psi$
    \STATE Update $\pi_\theta$ and $V_\psi$ with MAPPO
\ENDFOR
\STATE \RETURN $\theta$
\end{algorithmic}
\end{algorithm}

In Stage I, $\mathcal{R}_s$ is the topology-training reward. In Stage II, $\mathcal{R}_s$ is the topology-and-geometry co-design reward. In Stage III, $\mathcal{R}_s$ is the frontier-guided reward computed from the target-wise archive.

\section{Reinforcement Learning Formulation}
\label{app:rl_formulation}

\subsection{Decentralized partially observable Markov decision process}

We model structural-unit generation as a fully cooperative decentralized partially observable Markov decision process (Dec-POMDP)~\citep{bernstein2002complexity}. The process is denoted by
\begin{equation}
\mathcal{M}
=
\langle
\mathcal{N},
\mathcal{S},
\{\mathcal{O}_i\}_{i=1}^N,
\{\mathcal{A}_i\}_{i=1}^N,
P,
O,
r,
\gamma
\rangle .
\end{equation}
The tuple components are defined as follows:
\begin{itemize}
    \item $\mathcal{N}=\{1,\ldots,N\}$ is the set of cooperative agents;
    \item $\mathcal{S}$ is the global state space;
    \item $\mathcal{O}_i$ is the local observation space of agent $i$;
    \item $\mathcal{A}_i$ is the action space of agent $i$;
    \item $P(s'\mid s,\boldsymbol{a})$ is the transition function under joint action $\boldsymbol{a}=(a^1,\ldots,a^N)$;
    \item $O(o^1,\ldots,o^N\mid s',\boldsymbol{a})$ is the observation function;
    \item $r(s,\boldsymbol{a})$ is the shared reward function;
    \item $\gamma\in[0,1)$ is the discount factor.
\end{itemize}

At step $t$, let $\mathcal{I}_t\subseteq\mathcal{N}$ denote the active agents whose assigned unit segments have not been completed. Each active agent $i\in\mathcal{I}_t$ receives a local observation $o_t^i$ and samples an action from the shared actor:
\begin{equation}
a_t^i\sim\pi_\theta(\cdot\mid o_t^i).
\end{equation}
The joint policy factorizes over active agents as
\begin{equation}
\pi_\theta(\boldsymbol{a}_t\mid \boldsymbol{o}_t)
=
\prod_{i\in\mathcal{I}_t}
\pi_\theta(a_t^i\mid o_t^i),
\end{equation}
where $\boldsymbol{a}_t=\{a_t^i\}_{i\in\mathcal{I}_t}$. The environment writes the selected single-position or paired-unit assignments into the partial sequence and then advances the active cursors. The transition is deterministic given the sampled joint action. All active agents receive the same reward $r_t$.

The objective is to learn a shared policy that maximizes the expected discounted return:
\begin{equation}
J(\pi_\theta)
=
\mathbb{E}_{\pi_\theta}
\left[
\sum_{t=0}^{T-1}
\gamma^t r_t
\right].
\end{equation}
In \our, rewards are terminal at the end of sequence generation. Stage I uses the topology-training reward, Stage II uses the topology-and-geometry co-design reward, and Stage III uses the frontier-guided reward. The rollout dynamics remain the same across stages; only the conditioning information and terminal reward construction change.

The global design state includes the target topology, the ordered structural-unit list, the current partial sequence, the assignment mask, all agent cursors, and the rollout progress. In Stage II and Stage III, the state also includes target $\mathrm{3D}$ information. Each agent observes the unit at its cursor, the unit type, the current assignment state of that unit, and a local topology context. When $\mathrm{3D}$ conditioning is used, the observation is augmented with the current unit's $\mathrm{3D}$ embedding and a pooled local $\mathrm{3D}$ context. The episode ends when all structural units are assigned.

\subsection{Centralized training and decentralized execution}

\our\ uses centralized training and decentralized execution (CTDE), a standard paradigm in cooperative MARL~\citep{lowe2017multi,hu2026beyond,sunehag2018value,cui2026qmip,rashid2018qmix,tang2026hmu,foerster2018counterfactual,tang2026autonomous,yu2022surprising}. The actor is shared across agents and uses unit-type-specific output heads for single-position and paired units. During execution, each agent acts only from its local observation. During training, a centralized critic receives a global critic state $s_t^{\mathrm{crit}}$ constructed from the partial design state and outputs $V_\phi(s_t^{\mathrm{crit}})$. This separation matches the RNA design problem: actions are local to structural units, but sequence quality is evaluated by a global structural objective.

For a rollout, we compute temporal-difference residuals and generalized advantage estimates (GAE)~\citep{schulman2015high} as
\begin{equation}
\delta_t
=
r_t
+
\gamma V_\phi(s_{t+1}^{\mathrm{crit}})
-
V_\phi(s_t^{\mathrm{crit}}),
\qquad
A_t
=
\sum_{\ell=0}^{T-t-1}
(\gamma\lambda_{\mathrm{GAE}})^\ell
\delta_{t+\ell}.
\end{equation}
Because the reward is shared, the same advantage $A_t$ is assigned to all active agent transitions at step $t$.

\subsection{MAPPO objective}

Policy updates use the clipped MAPPO objective~\citep{yu2022surprising,schulman2017proximal}. Let $\theta_{\mathrm{old}}$ denote the policy parameters before the update. For an active agent transition, the probability ratio is
\begin{equation}
\rho_t^i(\theta)
=
\frac{
\pi_\theta(a_t^i\mid o_t^i)
}{
\pi_{\theta_{\mathrm{old}}}(a_t^i\mid o_t^i)
}.
\end{equation}
The actor objective is
\begin{equation}
\mathcal{L}_{\mathrm{actor}}(\theta)
=
\mathbb{E}_{t,i\in\mathcal{I}_t}
\left[
\min
\left(
\rho_t^i(\theta)A_t,
\operatorname{clip}(\rho_t^i(\theta),1-\epsilon,1+\epsilon)A_t
\right)
\right].
\end{equation}
The critic is trained to predict the target return $\hat{R}_t$:
\begin{equation}
\mathcal{L}_{\mathrm{value}}(\phi)
=
\mathbb{E}_t
\left[
\left(
V_\phi(s_t^{\mathrm{crit}})
-
\hat{R}_t
\right)^2
\right].
\end{equation}
The total loss minimized during training is
\begin{equation}
\mathcal{L}(\theta,\phi)
=
-
\mathcal{L}_{\mathrm{actor}}(\theta)
+
c_v\mathcal{L}_{\mathrm{value}}(\phi)
-
c_H\mathcal{H}(\pi_\theta).
\end{equation}
The same MAPPO objective is used across topology-aware initialization, topology-and-geometry co-design, and frontier-guided fine-tuning.

\subsection{Effective horizon}

Let $M$ be the number of structural units in the target topology. A single-agent unit-by-unit policy requires $M$ environment steps to complete the sequence. In contrast, \our\ splits the ordered unit list into $N$ contiguous segments and lets agents act jointly. The rollout length is therefore the maximum segment length, approximately $\lceil M/N\rceil$ when the split is balanced. This shorter effective horizon reduces delayed rewards and improves temporal credit assignment, which is especially useful because structural rewards are obtained only after a complete sequence is generated.

\section{Additional Results}
\label{app:additional_results}

In this section, we present additional experiments under various settings to further demonstrate the effectiveness and robustness of \our. Specifically, we include:
\begin{itemize}
    \item sensitivity analysis of the success thresholds;
    \item comparison between Top-10 sampling and frontier-guided fine-tuning;
    \item comparison between frontier-guided fine-tuning and fixed weighted rewards;
    \item ablation of the cooperative multi-agent formulation.
\end{itemize}

\subsection{Threshold Sensitivity}
\label{app:threshold_sensitivity_results}

\begin{figure}[!tbp]
\centering
\includegraphics[width=\linewidth]{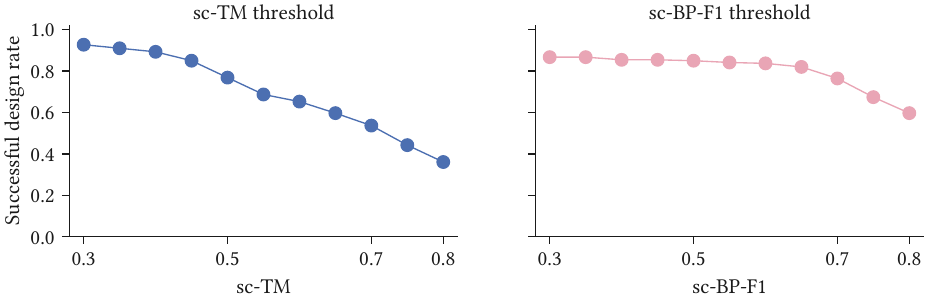}
\caption{
Sensitivity of \our\ to the joint success criterion.
Left: successful design rate when varying the $\mathrm{sc\text{-}TM}$ threshold while fixing $\mathrm{sc\text{-}BP\text{-}F1}$ $>0.50$.
Right: successful design rate when varying the $\mathrm{sc\text{-}BP\text{-}F1}$ threshold while fixing $\mathrm{sc\text{-}TM}$ $>0.45$.
}
\label{fig:app_threshold_sensitivity}
\end{figure}

A $\mathrm{TM\text{-}score}$ above $0.45$ is commonly used to indicate that two RNA structures share the same global topology~\citep{zhang2022us}. We therefore use $\mathrm{sc\text{-}TM}$ $>0.45$ as the tertiary-structure threshold in the main experiments, together with $\mathrm{sc\text{-}BP\text{-}F1}$ $>0.50$ to require secondary-structure consistency. Figure~\ref{fig:app_threshold_sensitivity} further evaluates how sensitive the successful design rate is to these thresholds. As expected, the success rate decreases as either requirement becomes stricter. When the $\mathrm{sc\text{-}TM}$ threshold is increased from $0.30$ to $0.80$ with $\mathrm{sc\text{-}BP\text{-}F1}$ fixed at $0.50$, the successful design rate decreases by $61.3\%$ relative to the loosest threshold setting. When the $\mathrm{sc\text{-}BP\text{-}F1}$ threshold is increased from $0.30$ to $0.80$ with $\mathrm{sc\text{-}TM}$ fixed at $0.45$, the rate decreases by $31.0\%$. These results show that joint success is sensitive to the required levels of both tertiary-structure and secondary-structure fidelity. Nevertheless, \our\ retains a substantial successful design rate under stricter criteria, supporting its robustness in generating high-quality RNA designs.

\subsection{Top 10 Sampling versus Frontier-guided Fine-tuning}
\label{app:top10_sampling_results}

\begin{figure}[!tbp]
\centering
\includegraphics[width=\linewidth]{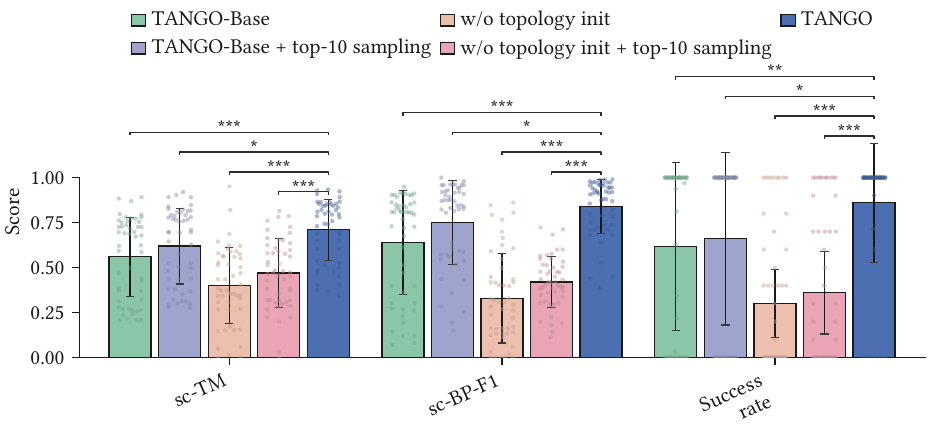}
\caption{
Comparison between Top-10 sampling and frontier-guided fine-tuning.
For each target, we sample sequences from 30 random seeds and select the 10 designs with the highest $\mathrm{sc\text{-}TM}$ for evaluation.
We compare raw policies, their Top-10 sampling variants, and the full \our\ model.
Bars show mean $\pm$ standard deviation across targets.
}
\label{fig:app_top10_sampling}
\end{figure}

Figure~\ref{fig:app_top10_sampling} compares frontier-guided fine-tuning with a Top-10 sampling strategy. We include \texttt{TANGO-Base}, an ablated variant without frontier-guided fine-tuning, and evaluate both its raw sampling performance and its Top-10 sampling performance. For each target, Top-10 sampling selects the 10 designs with the highest $\mathrm{sc\text{-}TM}$ from 30 sampled sequences. We also evaluate Top-10 sampling for the variant without topology-aware initialization.

The full \our\ model clearly outperforms all Top-10 sampling variants. Compared with \texttt{TANGO-Base} with Top-10 sampling, \our\ achieves relative improvements of $14.5\%$ in $\mathrm{sc\text{-}TM}$ and $12.0\%$ in $\mathrm{sc\text{-}BP\text{-}F1}$. Compared with the Top-10 sampling variant without topology-aware initialization, \our\ achieves relative improvements of $51.1\%$ in $\mathrm{sc\text{-}TM}$ and $100.0\%$ in $\mathrm{sc\text{-}BP\text{-}F1}$. This shows that the gains from \our\ cannot be explained by selecting better samples from a fixed policy.

Without policy optimization, Top-10 sampling still improves the raw policies, indicating that the sampled designs remain diverse enough for post-hoc selection to find stronger candidates. For \texttt{TANGO-Base}, Top-10 sampling improves $\mathrm{sc\text{-}TM}$ by $10.7\%$ and $\mathrm{sc\text{-}BP\text{-}F1}$ by $17.2\%$ over raw sampling. For the variant without topology-aware initialization, it improves $\mathrm{sc\text{-}TM}$ by $17.5\%$ and $\mathrm{sc\text{-}BP\text{-}F1}$ by $27.3\%$. These results show that sampling and reranking can improve candidate selection, but frontier-guided fine-tuning is more effective because it improves the policy to generate higher-quality designs rather than only selecting from existing samples.

\subsection{Frontier-guided Fine-tuning versus Weighted Rewards}
\label{app:weighted_reward_results}

\begin{figure}[!tbp]
\centering
\includegraphics[width=\linewidth]{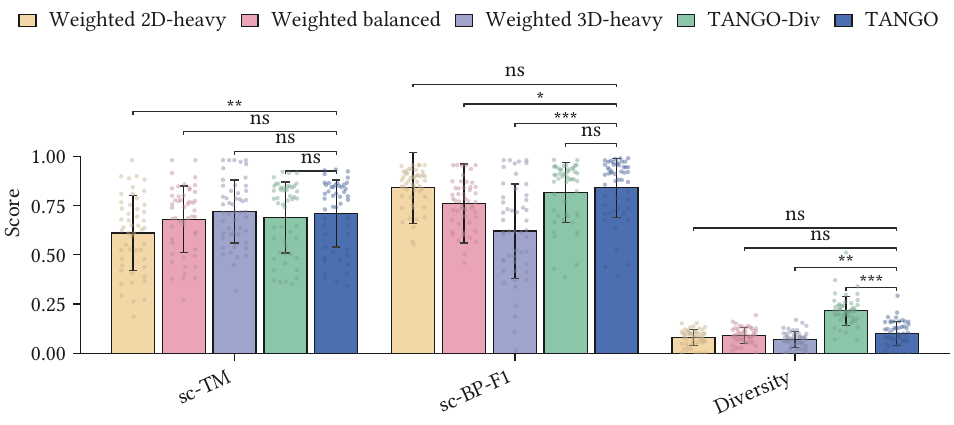}
\caption{
Comparison between frontier-guided fine-tuning and fixed weighted reward alternatives. Three scalarized reward settings with different $\mathrm{2D}$/$\mathrm{3D}$ weights are compared with \our\ and \texttt{TANGO-Div} on $\mathrm{sc\text{-}TM}$, $\mathrm{sc\text{-}BP\text{-}F1}$, and sequence diversity. 
Bars show mean $\pm$ standard deviation across targets. Significance markers denote \texttt{*} for $p<0.05$, \texttt{**} for $p<0.01$, \texttt{***} for $p<0.001$, and \texttt{ns} for not significant.
}
\label{fig:app_weighted_reward_bars}
\end{figure}

\begin{figure}[!tbp]
\centering
\includegraphics[width=0.60\linewidth]{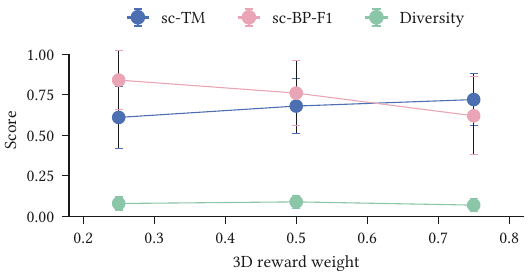}
\caption{
Trade-offs among multiple objectives under fixed weighted rewards. Changing the scalarization weights shifts the balance among $\mathrm{2D}$ fidelity, $\mathrm{3D}$ fidelity, and diversity, whereas frontier-guided fine-tuning directly optimizes the target-wise trade-off frontier.
}
\label{fig:app_weighted_reward_tradeoff}
\end{figure}

Figures~\ref{fig:app_weighted_reward_bars} and~\ref{fig:app_weighted_reward_tradeoff} compare frontier-guided fine-tuning with fixed weighted reward alternatives. The $\mathrm{2D}$-heavy setting matches \our\ in $\mathrm{sc\text{-}BP\text{-}F1}$ ($0.84$), but achieves lower $\mathrm{sc\text{-}TM}$ ($0.61$) and diversity ($0.08$). Conversely, the $\mathrm{3D}$-heavy setting achieves slightly higher $\mathrm{sc\text{-}TM}$ ($0.72$), but reduces $\mathrm{sc\text{-}BP\text{-}F1}$ to $0.62$. These results show that fixed scalarization can bias optimization toward one structural objective at the expense of others. In contrast, \our\ maintains a stronger balance between topology and geometry, while \texttt{TANGO-Div} further increases diversity from $0.10$ to $0.22$ with only small decreases in $\mathrm{sc\text{-}TM}$ and $\mathrm{sc\text{-}BP\text{-}F1}$. This supports the use of frontier-guided fine-tuning for target-wise multi-objective refinement without relying on manually chosen scalarization weights.

\subsection{Effect of the Cooperative Multi-agent Formulation}
\label{app:multi_agent_results}

\begin{figure}[!tbp]
\centering
\includegraphics[width=0.65\linewidth]{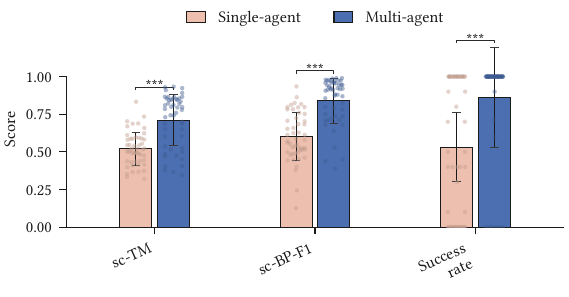}
\caption{
Ablation of the cooperative multi-agent formulation. 
We compare the default multi-agent TANGO with a single-agent variant under the same training and evaluation protocol. 
Bars show mean $\pm$ standard deviation across held-out targets. 
Significance markers denote \texttt{*} for $p<0.05$, \texttt{**} for $p<0.01$, \texttt{***} for $p<0.001$, and \texttt{ns} for not significant.
}
\label{fig:app_multi_agent_rl}
\end{figure}

Figure~\ref{fig:app_multi_agent_rl} evaluates the effect of the cooperative multi-agent formulation by comparing \our\ with a single-agent variant. The multi-agent formulation consistently improves all metrics, increasing $\mathrm{sc\text{-}TM}$ from $0.52$ to $0.71$, $\mathrm{sc\text{-}BP\text{-}F1}$ from $0.60$ to $0.84$, and success rate from $0.53$ to $0.86$. These correspond to relative improvements of $36.5\%$, $40.0\%$, and $62.3\%$, respectively. The results show that decomposing RNA sequence construction into coordinated local decisions benefits both tertiary-structure recovery and secondary-structure consistency.

\section{Model Architecture}
\label{app:architecture}

\subsection{Topology and Unit Encoders}

The policy operates on structural units, while the first representation is constructed at the residue level. For residue $p$, the topology encoder embeds the dot-bracket token $b_p$ and concatenates it with structural metadata $\boldsymbol{m}_p$, including pairing status, normalized position, pairing distance, and pseudoknot-related indicators. The token vocabulary covers unpaired positions, standard parenthesis brackets, and additional bracket symbols used for pseudoknotted base pairs. The concatenated representation is mapped by an MLP to a residue-level topology embedding $\boldsymbol{h}^{\mathcal{T}}_p\in\mathbb{R}^{256}$.

The unit encoder then converts residue embeddings into the structural-unit representation used by all agents. For a single-position unit $u_j=\{p\}$, the input is $\boldsymbol{h}^{\mathcal{T}}_p$ together with unit metadata $\boldsymbol{m}_{u_j}$, such as normalized position and local topology context. For a paired unit $u_j=\{p,q\}$, the input contains $\boldsymbol{h}^{\mathcal{T}}_p$, $\boldsymbol{h}^{\mathcal{T}}_q$, their elementwise product, their absolute difference, and pair-level metadata such as normalized pair distance, normalized endpoints, and pseudoknot/context indicators. Separate MLPs encode single and paired units into a shared $256$-dimensional unit space. This design aligns the representation with the action granularity: paired residues are represented and assigned as one coupled decision rather than two independent nucleotide decisions.

A local topology context vector is computed for the current unit by mean pooling unit embeddings in a fixed local window around that unit. This context is used as local information for the actor. The centralized critic receives a separate global summary of the partial design state.

\subsection{Shared Actor with Unit-specific Heads}

All agents share one actor. At a decision step, the actor input for an active agent is the current unit embedding, the assignment state of that unit, and the local topology context, corresponding to the observation $o_t^i=[\boldsymbol{e}_{j_t^i},\boldsymbol{q}_{t,j_t^i},\boldsymbol{c}_{j_t^i}]$ used in the main method. A shared MLP trunk maps this observation to a hidden representation.

The final action distribution is unit-specific. The single-unit head produces logits over $\{\texttt{A},\texttt{U},\texttt{G},\texttt{C}\}$, and the paired-unit head produces logits over $\{\texttt{AU},\texttt{UA},\texttt{GC},\texttt{CG},\texttt{GU},\texttt{UG}\}$. Agents with no remaining units are inactive at that step and do not contribute policy transitions.

\subsection{Centralized Critic}

The critic is centralized during training. In the topology-aware initialization stage, it receives a pooled representation of the global partial design state, summarizing the target topology, current partial sequence, assignment mask, and rollout progress. In the topology-and-geometry co-design and frontier-guided fine-tuning stages, the critic additionally receives a pooled $\mathrm{3D}$ target condition. The actor remains decentralized: at inference time, each active agent only uses its local observation and the shared actor parameters.

\subsection{\texorpdfstring{$\mathrm{3D}$}{3D} Target Encoder}

The target tertiary structure is represented with three beads per nucleotide: $\mathrm{P}$, $\mathrm{C4}^{\prime}$, and the glycosidic nitrogen atom, $\mathrm{N1}$ for pyrimidines and $\mathrm{N9}$ for purines~\citep{dawson2016coarse}. This coarse-grained representation preserves backbone geometry and approximate base placement while avoiding all-atom inputs.

\our\ uses a GVP-GNN encoder to map the target geometry to residue-level embeddings~\citep{jing2021learning,joshi2025grnade}. The encoder constructs a spatial graph whose nodes are nucleotides and whose edges connect nearby residues in $\mathrm{3D}$ space. Node and edge features contain scalar channels, such as distances and positional encodings, and vector channels, such as backbone directions and relative displacement vectors. Native nucleotide identities are not used as input features, preventing leakage from the target sequence. A GVP-GNN layer updates scalar-vector node features by aggregating messages from spatial neighbors:
\begin{equation}
\begin{aligned}
(\boldsymbol{m}_i^{(l)},\vec{\boldsymbol{m}}_i^{(l)})
&=
\operatorname{AGG}_{j\in\mathcal{N}(i)}
\phi_{\mathrm{msg}}^{(l)}
\left(
(\boldsymbol{s}_i^{(l-1)},\vec{\boldsymbol{v}}_i^{(l-1)}),
(\boldsymbol{s}_j^{(l-1)},\vec{\boldsymbol{v}}_j^{(l-1)}),
(\boldsymbol{e}_{ij},\vec{\boldsymbol{e}}_{ij})
\right),\\
(\boldsymbol{s}_i^{(l)},\vec{\boldsymbol{v}}_i^{(l)})
&=
\phi_{\mathrm{upd}}^{(l)}
\left(
(\boldsymbol{s}_i^{(l-1)},\vec{\boldsymbol{v}}_i^{(l-1)}),
(\boldsymbol{m}_i^{(l)},\vec{\boldsymbol{m}}_i^{(l)})
\right).
\end{aligned}
\end{equation}
Here $\phi_{\mathrm{msg}}^{(l)}$ and $\phi_{\mathrm{upd}}^{(l)}$ are GVP-based message and update functions. Scalar channels provide rotation-invariant quantities for the downstream policy, while vector channels allow the encoder to propagate orientation-dependent geometric information during message passing. The resulting residue-level geometric embeddings are used as conditioning signals for the structural-unit actor.

Residue-level geometric embeddings are pooled to match the structural units. A single-position unit inherits the embedding of its residue. A paired unit uses the mean of the two residue embeddings. We also compute a unit center from the target $\mathrm{C4}^{\prime}$ coordinates and build a unit-level $\mathrm{3D}$ nearest-neighbor list. The local $\mathrm{3D}$ context for a unit is the mean of the embeddings of its nearest neighboring units, using $15$ nearest units in the reported configuration.

\subsection{Zero-initialized Geometry Adapters}

The topology-and-geometry co-design stage initializes from the topology-aware actor and adds two $\mathrm{3D}$-conditioned paths. For unit $u_j$, let $\boldsymbol{g}_j$ denote the unit-level $\mathrm{3D}$ embedding and $\boldsymbol{c}^{\mathrm{3D}}_j$ denote the local $\mathrm{3D}$ context. We define the geometry condition as $\boldsymbol{d}^{\mathrm{3D}}_j=[\boldsymbol{g}_j,\boldsymbol{c}^{\mathrm{3D}}_j]$.

The first $\mathrm{3D}$-conditioned path applies FiLM modulation to the inherited hidden state:
\begin{equation}
\widetilde{\boldsymbol{z}}_j
=
\left(1+\gamma(\boldsymbol{d}^{\mathrm{3D}}_j)\right)\odot \boldsymbol{z}_j
+
\beta(\boldsymbol{d}^{\mathrm{3D}}_j),
\end{equation}
where $\boldsymbol{z}_j$ is the hidden representation produced by the topology-aware base path. The second path directly maps the topology-aware unit state together with the geometry condition through a small $\mathrm{3D}$-conditioned network. The final action logits are
\begin{equation}
\ell_j
=
\ell_j^{\mathrm{base}}
+
\Delta \ell_j^{\mathrm{FiLM}}
+
\Delta \ell_j^{\mathrm{3D}}.
\end{equation}
The last linear layers of both $\mathrm{3D}$-conditioned paths are initialized to zero. Therefore, at the beginning of topology-and-geometry co-design, the transferred policy is close to the topology-aware policy. Geometry conditioning can then gradually adjust topology-compatible sequence decisions toward the target tertiary structure.

\section{Training Details}
\label{app:training_details}

The three learning stages share the structural-unit rollout interface and the MAPPO update described in Appendix~\ref{app:rl_formulation}. At the beginning of each episode, the target is parsed into ordered structural units, and cooperative agents fill these units until a complete candidate sequence is generated. The learning signal is assigned at the end of the episode. Stage I uses a topology-training reward, Stage II uses a topology-and-geometry co-design reward, and Stage III uses a frontier-guided reward based on target-wise multi-objective improvement.

Across stages, MAPPO uses discount factor $\gamma=0.99$, GAE parameter $\lambda_{\mathrm{GAE}}=0.95$, PPO clip range $0.2$, value coefficient $0.5$, and gradient clipping at norm $5.0$. Stage-specific rollout, optimization, and reward settings are summarized below. Protocol choices such as the fold-back oracle, structure-alignment backend, and frontier objective set are described in prose when they define the training interface rather than scalar hyperparameters.

\subsection{Stage I Hyperparameters}

Stage I trains the topology-aware policy on secondary-structure targets. After the agents generate a complete sequence, the secondary-structure oracle evaluates its agreement with the target secondary structure, and the resulting topology-training reward is shared by all agents. This stage teaches the policy to make base-pair-compatible structural-unit assignments before tertiary-structure conditioning is introduced. Training is performed on a server with one \texttt{NVIDIA} \texttt{H100} GPU. The training takes approximately $5$ hours, and inference for a single sample takes only a few seconds. The rollout and optimization settings are shown in Table~\ref{tab:app_stage1_rollout_hparams}.

\begin{table}[!tbp]
\centering
\small
\renewcommand{\arraystretch}{1.2}
\caption{Stage I rollout and optimization settings.}
\label{tab:app_stage1_rollout_hparams}
\begin{tabular}{@{}ll@{}}
\toprule
\textbf{Setting} & \textbf{Value} \\
\midrule
Number of agents & 4 \\
Actor learning rate & $1\times 10^{-4}$ \\
Critic learning rate & $1\times 10^{-4}$ \\
Encoder learning-rate scale & 0.3 \\
Optimizer & AdamW \\
Weight decay & $1\times 10^{-4}$ \\
Learning-rate decay floor & 0.3 \\
Buffer size & 4096 \\
Epochs per update & 4 \\
Minibatches per update & 4 \\
Entropy coefficient & 0.01 \\
\bottomrule
\end{tabular}
\end{table}

\subsection{Stage II Hyperparameters}

Stage II starts from the topology-aware policy and trains a topology-and-geometry co-design policy. For each completed sequence, we predict its tertiary structure with the fold-back oracle, align it to the target structure, and compute the $\mathrm{TM\text{-}score}$ as the $\mathrm{3D}$ reward. The predicted tertiary structure is also annotated into a secondary structure, which is used to compute the $\mathrm{BP\text{-}F1}$ topology reward. Following the main text, the Stage II co-design reward contains three terms: a $\mathrm{3D}$ fidelity term, a $\mathrm{2D}$ fidelity term, and a hinge penalty when the $\mathrm{2D}$ fidelity falls below the target threshold:
\begin{equation}
R_{\mathrm{co}}(x_T;\mathcal{G}^\star,\mathcal{T}^\star)
=
R_{\mathrm{3D}}(x_T;\mathcal{G}^\star)
+
\lambda_{\mathrm{2D}}R_{\mathrm{2D}}(x_T;\mathcal{T}^\star)
-
\lambda_{\mathrm{viol}}
\max
\left(
0,
\tau_{\mathrm{2D}}
-
R_{\mathrm{2D}}(x_T;\mathcal{T}^\star)
\right).
\end{equation}
Here $R_{\mathrm{3D}}$ is measured by $\mathrm{TM\text{-}score}$, and $R_{\mathrm{2D}}$ is measured by $\mathrm{BP\text{-}F1}$. $\mathrm{RMSD}$ is recorded for evaluation but is not used as a reward term. The rollout, optimization, and reward settings are shown in Table~\ref{tab:app_stage2_rollout_hparams}.

\begin{table}[!tbp]
\centering
\small
\renewcommand{\arraystretch}{1.2}
\caption{Stage II rollout, optimization, and co-design reward settings.}
\label{tab:app_stage2_rollout_hparams}
\begin{tabular}{@{}ll@{}}
\toprule
\textbf{Setting} & \textbf{Value} \\
\midrule
Number of agents & 4 \\
Actor learning rate & $5\times 10^{-4}$ \\
Critic learning rate & $5\times 10^{-4}$ \\
Learning-rate decay floor & 0.3 \\
Buffer size & 1024 \\
Epochs per update & 10 \\
Minibatches per update & 2 \\
Entropy coefficient & 0.01 \\
\midrule
$\lambda_{\mathrm{2D}}$ & 1.0 \\
$\lambda_{\mathrm{viol}}$ & 2.0 \\
$\tau_{\mathrm{2D}}$ & 0.4 \\
\bottomrule
\end{tabular}
\end{table}

\subsection{Stage III Hyperparameters}

Stage III applies frontier-guided fine-tuning. For each target, we warm-start an archive with candidate sequences sampled from the co-design policy. During fine-tuning, each update samples complete candidate sequences, evaluates them with the same fold-back protocol, and assigns each trajectory a terminal reward given by its exclusive hypervolume contribution to the target-wise frontier. The default bi-objective frontier uses $\mathrm{sc\text{-}BP\text{-}F1}$ and $\mathrm{sc\text{-}TM}$, while the diversity-aware variant additionally includes sequence diversity. The fine-tuning settings are shown in Table~\ref{tab:app_ft_hparams}.

\begin{table}[!tbp]
\centering
\small
\renewcommand{\arraystretch}{1.2}
\caption{Stage III frontier-guided fine-tuning settings.}
\label{tab:app_ft_hparams}
\begin{tabular}{@{}ll@{}}
\toprule
\textbf{Setting} & \textbf{Value} \\
\midrule
Warm-start archive size & 64 \\
Actor learning rate & $1\times 10^{-4}$ \\
Critic learning rate & $1\times 10^{-4}$ \\
Buffer size & 1024 \\
Epochs per update & 4 \\
Minibatches per update & 2 \\
Entropy coefficient & 0.02 \\
\bottomrule
\end{tabular}
\end{table}

\section{Evaluation Metrics}
\label{app:metrics}

\subsection{Fold-back Self-consistency Protocol}

The main evaluation follows a fold-back self-consistency protocol. Let $F_{\mathrm{3D}}$ denote the selected tertiary-structure prediction oracle, and let $A_{\mathrm{2D}}$ denote the $\mathrm{3D}$-to-$\mathrm{2D}$ annotation procedure. For a target with native sequence $x^\star$, we define the reference tertiary structure and its associated secondary topology as
\begin{equation}
\mathcal{G}^\star
=
F_{\mathrm{3D}}(x^\star),
\qquad
\mathcal{T}^\star
=
A_{\mathrm{2D}}(\mathcal{G}^\star).
\end{equation}
For a designed sequence $x$, we compute its fold-back tertiary structure and annotated secondary topology as
\begin{equation}
\widehat{\mathcal{G}}(x)
=
F_{\mathrm{3D}}(x),
\qquad
\widehat{\mathcal{T}}^{\mathrm{ann}}(x)
=
A_{\mathrm{2D}}\left(\widehat{\mathcal{G}}(x)\right).
\end{equation}
The designed sequence is evaluated by comparing $\widehat{\mathcal{G}}(x)$ with $\mathcal{G}^\star$ and $\widehat{\mathcal{T}}^{\mathrm{ann}}(x)$ with $\mathcal{T}^\star$. This differs from directly comparing a predicted design structure with experimental coordinates. Since the reference and designed sequences are folded by the same oracle, the shared predictor-specific inductive bias is expected to cancel out in the structural similarity calculation~\citep{gao2024proteininvbench}. We additionally report cross-oracle evaluation of \our\ designs in Figure~\ref{fig:cross_oracle} to examine whether the observed trends remain consistent under an independent fold-back predictor.

\subsection{\texorpdfstring{$\mathrm{3D}$}{3D} Self-consistency Metrics}

We report tertiary-structure fidelity using $\mathrm{sc\text{-}TM}$ and $\mathrm{sc\text{-}RMSD}$. The $\mathrm{sc\text{-}TM}$ score is the $\mathrm{TM\text{-}score}$ computed by USalign between the designed fold-back structure and the reference fold-back structure:
\begin{equation}
\mathrm{sc\text{-}TM}(x,x^\star)
=
\operatorname{TM}
\left(
\widehat{\mathcal{G}}(x),
\mathcal{G}^\star
\right).
\end{equation}
$\mathrm{TM\text{-}score}$ is length-normalized and evaluates global fold similarity~\citep{zhang2022us}. Given an aligned set of positions $\mathcal{A}$ with pairwise distances $d_i$, USalign computes $\mathrm{TM\text{-}score}$ using its standard length normalization:
\begin{equation}
\operatorname{TM}
=
\max
\left[
\frac{1}{L_{\mathrm{ref}}}
\sum_{i\in\mathcal{A}}
\frac{1}{1+\left(d_i/d_0(L_{\mathrm{ref}})\right)^2}
\right].
\end{equation}
The $\mathrm{sc\text{-}RMSD}$ score is the $\mathrm{RMSD}$ over the aligned region after structural alignment:
\begin{equation}
\mathrm{sc\text{-}RMSD}
=
\sqrt{
\frac{1}{|\mathcal{A}|}
\sum_{i\in\mathcal{A}} d_i^2
}.
\end{equation}
Thus, $\mathrm{sc\text{-}TM}$ measures global fold agreement, whereas $\mathrm{sc\text{-}RMSD}$ reflects geometric deviation within the aligned region.

\subsection{\texorpdfstring{$\mathrm{2D}$}{2D} Self-consistency Metrics}

Secondary-topology fidelity is measured by comparing the target topology $\mathcal{T}^\star$ with the annotated topology $\widehat{\mathcal{T}}^{\mathrm{ann}}(x)$ extracted from the designed fold-back structure. Let $\mathcal{P}^\star$ denote the target base-pair set and let $\widehat{\mathcal{P}}(x)$ denote the base-pair set extracted from $\widehat{\mathcal{T}}^{\mathrm{ann}}(x)$. The self-consistent base-pair F1 score is
\begin{equation}
\mathrm{sc\text{-}BP\text{-}F1}(x,x^\star)
=
\frac{
2|\widehat{\mathcal{P}}(x)\cap\mathcal{P}^\star|
}{
|\widehat{\mathcal{P}}(x)|+|\mathcal{P}^\star|
}.
\end{equation}
When both pair sets are empty, we define $\mathrm{sc\text{-}BP\text{-}F1}=1$; when only one pair set is empty, the score is $0$.

\subsection{Target-level Aggregation}

Each method generates multiple designs for each target. We first aggregate metrics within each target and then report averages across test targets. For $\mathrm{sc\text{-}TM}$, $\mathrm{sc\text{-}RMSD}$, and $\mathrm{sc\text{-}BP\text{-}F1}$, we compute the target-level mean over sampled designs and then average over targets. A design is considered successful if it satisfies
\begin{equation}
\mathrm{sc\text{-}TM}(x,x^\star)\ge \tau_{\mathrm{3D}},
\qquad
\mathrm{sc\text{-}BP\text{-}F1}(x,x^\star)\ge \tau_{\mathrm{2D}}.
\end{equation}
The success rate reports the average fraction of sampled designs satisfying both thresholds. The solved target rate reports the fraction of test targets for which at least one sampled design satisfies both thresholds. Appendix~\ref{app:threshold_sensitivity_results} evaluates how \our\ changes under alternative $\mathrm{sc\text{-}TM}$ and $\mathrm{sc\text{-}BP\text{-}F1}$ thresholds. This target-level aggregation prevents targets with more valid sampled outputs from dominating the benchmark average.

\subsection{Novelty and Diversity}

Sequence novelty is defined as the normalized Hamming distance between a designed sequence and the native sequence of the target:
\begin{equation}
\operatorname{Novelty}(x,x^\star)
=
\frac{1}{L}
\sum_{p=1}^{L}
\mathbb{I}[x_p\ne x^\star_p].
\end{equation}
For a target-level set of designed sequences $\mathcal{X}_t=\{x^{(1)},\ldots,x^{(m)}\}$ with common length $L$, sequence diversity is the average pairwise normalized Hamming distance:
\begin{equation}
\operatorname{Diversity}(\mathcal{X}_t)
=
\frac{2}{m(m-1)}
\sum_{1\le i<j\le m}
\frac{1}{L}
\sum_{p=1}^{L}
\mathbb{I}\left[x^{(i)}_p\ne x^{(j)}_p\right],
\qquad m\ge 2.
\end{equation}
When diversity is analyzed under structural success, $\mathcal{X}_t$ is restricted to successful designs for that target, so that random failed sequences do not artificially inflate diversity.

\section{Limitations and Broader Impact}

\paragraph{Future Work and Limitations}

Future work could investigate model merging~\citep{wang2026separatingcommonuniquedirections} across policies adapted from the same Stage-II initialization, assessing whether a merged policy can retain useful design capabilities across RNA targets. Future evaluation could also consider dimensions beyond structural fidelity, including measures of biological function~\citep{huang2025graphfla}.

Beyond the design method itself, structure predictors are a key component of the RNA design loop. The design framework is not tied to a specific structure-prediction oracle, but more accurate predictors can improve the translation of computational designs into experimentally useful molecules. Recent progress in RNA and biomolecular structure prediction has made these predictors increasingly accurate, and they have already been used as practical components in experimental RNA engineering pipelines~\citep{lim2026programmable,robson2025toehold}. Future advances in RNA structure prediction are expected to further improve the reliability and practical utility of \our.

\paragraph{Broader Impact}

\our\ may contribute to synthetic biology, RNA-based therapeutics, and studies of gene regulation by improving the computational design of RNAs with desired structural properties. Such methods could support the engineering of functional RNA circuits, aptamers, and ribozymes. However, AI-driven biological sequence design also raises concerns about reliability, interpretability, and experimental validation. Recent work on genomic models advocates systematic evaluation of explanations for consistency, faithfulness, and biological validity~\citep{zhou2026genomic}. Such assessment will also be important when using model interpretations to guide RNA engineering.

\clearpage
\endgroup

\end{document}